\documentclass[accepted]{uai2026} 
                        
\usepackage[american]{babel}
\usepackage{amsmath}
\usepackage{amssymb}
\usepackage{mathtools}
\usepackage{amsthm}
\usepackage{multirow}
\usepackage{multicol}
\usepackage{color}
\usepackage{caption}
\usepackage{booktabs}
\usepackage{multirow}
\usepackage{algorithm}
\usepackage{algorithmic}
\usepackage{setspace}
\usepackage{pifont}
\usepackage{colortbl}
\usepackage{subcaption}
\usepackage{natbib} 
\usepackage{mathtools} 
\usepackage{booktabs} 
\usepackage{tikz} 

\title{Set-based v.s. Distribution-based Representations of Epistemic Uncertainty:\\  A Comparative Study}

\author[1,2]{{Kaizheng Wang}\thanks{This work was initiated at KU Leuven and primarily completed at Nanyang Technological University. Corresponding author: Kaizheng Wang (\texttt{kaizheng.wang@ntu.edu.sg}).}}
\author[2,5]{Yunjia Wang}
\author[3]{Fabio Cuzzolin}
\author[4,5]{David Moens}
\author[2]{Hans Hallez}
\author[1]{Siu Lun Chau}

\affil[1]{%
	College of Computing and Data Science\\
	Nanyang Technological University\\
	Singapore
}  
\affil[2]{%
Department of Computer Science\\
KU Leuven\\
Belgium 
}
\affil[3]{%
School of Engineering, Computing, and Mathematics\\
Oxford Brookes University\\ U.K.
}
\affil[4]{%
Department of Mechanical Engineering\\
KU Leuven\\
Belgium
 }
\affil[5]{%
Flanders Make@KU Leuven\\
Belgium
}
\begin{document}
\maketitle

\begin{abstract}
Epistemic uncertainty in neural networks is commonly modeled using two second-order paradigms: distribution-based representations, which rely on posterior parameter distributions, and set-based representations based on credal sets. These frameworks are often regarded as fundamentally non-comparable due to differing semantics, assumptions, and evaluation practices, leaving their relative merits unclear. Empirical comparisons are further confounded by variations in the underlying predictive models. To clarify this issue, we present a controlled comparative study enabling principled, like-for-like evaluation of the two paradigms. Both representations are constructed from the same finite collection of predictive distributions generated by a shared neural network, isolating representational effects from predictive accuracy. Our study evaluates each representation through the lens of $3$ uncertainty measures across $14$ benchmarks, including selective prediction and out-of-distribution detection, spanning $6$ underlying predictive models and $10$ independent runs per configuration. Our results show that meaningful comparison between these seemingly non-comparable frameworks is both feasible and informative, providing insights into how second-order representation choices impact practical uncertainty-aware performance.
\end{abstract}

\section{Introduction}
\label{Sec: Intro}
Recent research has increasingly emphasized the representation and quantification of epistemic uncertainty (EU) in neural networks (NNs) to improve the robustness and reliability, particularly in safety-critical settings~\citep{zhou2012learning, pmlr-v151-tuo22a, mukhoti2023deep, mehrtens2023benchmarking, chau2025integral}. EU captures a model’s incomplete knowledge of the true input–output relationship and reflects uncertainty that is, in principle, reducible with additional information. Modeling EU often requires a second-order formalism capable of expressing uncertainty over the model’s own probabilistic predictions~\citep{hullermeier2021aleatoric, WangTPAMI}.

Two dominant paradigms have emerged for representing such second-order uncertainty. The first adopts a distribution-based representation, where uncertainty is modeled via probability distributions over model parameters or predictions. This perspective underlies Bayesian neural networks (BNNs) as well as practical approximations such as deep ensembles (DE)~\citep{blundell2015weight, krueger2017Bayesian, lakshminarayanan2017simple}. In practice, it is often simply a uniform distribution placed over a finite collection of (sampled) predictive distributions. The second paradigm employs a set-based representation, in which uncertainty is encoded by sets of plausible predictive distributions, most commonly credal sets---convex sets of probability distributions---as used in recent credal classification frameworks~\citep{levi1980enterprise, wang2024CredalEnsembles, wang2026Distill, wang2025credalWrapper, lohr2025credal}.

Despite their shared objective of capturing epistemic uncertainty, these paradigms are frequently regarded as fundamentally non-comparable. The two frameworks differ in semantics, mathematical structure, modeling assumptions, and evaluation methodology. Distribution-based approaches are often interpreted through Bayesian lenses, whereas credal approaches adopt imprecision-aware or set-valued perspectives. Consequently, existing discussions of their relative merits remain largely conceptual, and empirical comparisons are difficult to interpret due to confounding factors. In particular, comparisons typically involve predictors derived from different learning algorithms, architectures, or training procedures, making it unclear whether observed differences arise from predictive accuracy, optimization effects, or the uncertainty representation itself.

This ambiguity raises a central yet unresolved question: how do second-order representation choices themselves influence practical uncertainty-aware behavior? Addressing this question requires isolating representational effects from predictive ones---a requirement rarely satisfied in existing studies. Prior empirical work predominantly focuses on within-paradigm comparisons~\citep{mehrtens2023benchmarking, lohr2025credal} or evaluates representations using a single uncertainty metric or downstream task~\citep{sale2024secondorder, chau2025integral}, limiting the ability to draw more general conclusions. As a result, the practical implications of choosing between distributional and set-based second-order representations remain poorly understood.

\begin{figure}[t]
\centering
\includegraphics[width=\linewidth]{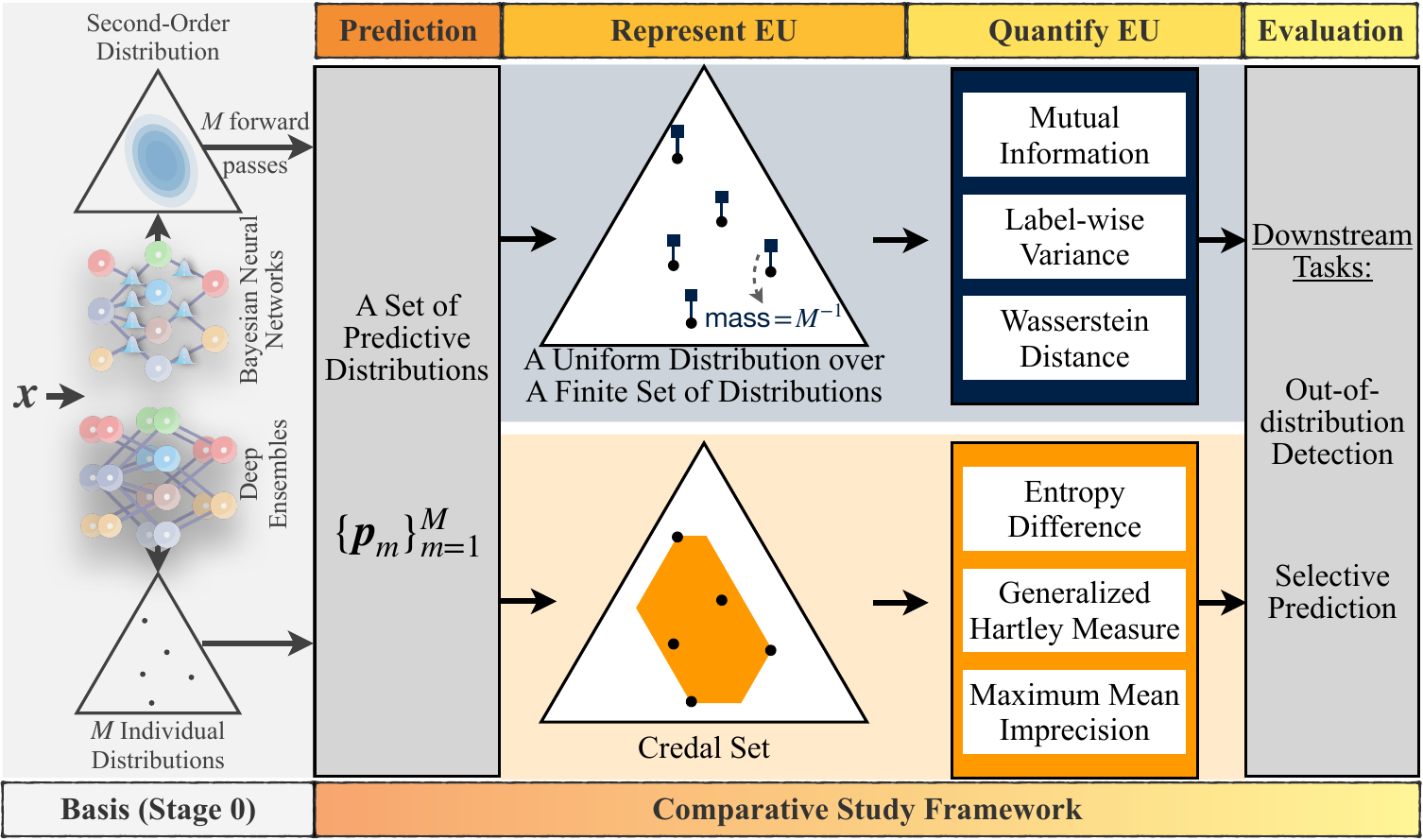}
\caption{Illustration of our comparative study framework.}
\label{FIG: Overview}
\end{figure}

\textbf{Novelty and contributions.} 
To address this issue, we conduct a controlled comparative study in classification settings, providing a unified framework for evaluating different second-order uncertainty representations. The study, as summarized in \figurename~\ref{FIG: Overview}, is designed to ensure rigorous and like-for-like comparison by explicitly controlling for confounding factors.

\textbf{(1)}~To eliminate effects arising from differences in model and learning assumptions, both representations are constructed from a finite collection of predictive distributions generated by the same neural network (either a Bayesian neural network or an ensemble). Within this setting, the distribution-based representation is defined as a uniform distribution over the predictive set, while the credal representation is derived from the identical predictions via class-wise probability interval construction \citep{probability_interval_1994}. 
\textbf{(2)}~To reduce dependence on any single uncertainty metric, we evaluate each representation using multiple uncertainty measures and perform both intra- and inter-representation comparisons. 
\textbf{(3)}~Furthermore, to broaden the evaluation scenarios, experiments are conducted across multiple neural network architectures and widely used benchmarks, including selective prediction---where an EU-aware model abstains on samples with high EU estimates to reduce misclassification risk---and out-of-distribution (OOD) detection.

\textbf{Three key findings} emerge from our study. First, no representation exhibits uniform superiority independent of the associated uncertainty measure; conclusions depend critically on the interaction between representation and metric. Second, out-of-distribution detection more clearly exposes representational differences than selective prediction. Third, reliable uncertainty quantification depends jointly on representation and uncertainty measure, with different metrics yielding substantially different behavior even within the same representation. See Section~\ref{Sec: ResultsAnalysis} for how our results support these analyses.

\textbf{Paper outline.}
The remainder of this paper is organized as follows. Section~\ref{Sec: preliminaries} introduces the preliminaries about uncertainty representation, quantification, and evaluation, respectively. Section~\ref{Sec: Experiment} and Section~\ref{Sec: ResultsAnalysis} present the experimental setups and the comparative analysis in full detail, respectively. Section~\ref{Sec: DiscussConclude} concludes this work with discussions. 

\textbf{Further related work.} Alternative second-order representations in classification include Dirichlet-based models \citep{malinin2018predictive, charpentier2020posterior} and random sets \citep{manchingal2025randomset}. However, these frameworks do not provide a principled mechanism for construction from general EU-aware predictors or systematic translation across representations, making controlled, like-for-like comparisons difficult. We therefore restrict our study to representations that can be consistently derived from a shared set of predictive distributions. Prior efforts toward cross-representation evaluation remain limited. \citet{manchingal2025frame} converted Bayesian neural network and deep ensemble predictions into credal sets, focusing on the accuracy–precision trade-off for model selection. \citet{mucsnyi2024benchmarking} examined uncertainty disentanglement but did not consider credal representations. Recent uncertainty measures \citep{lohr2025credal, chau2026quantifying} have similarly concentrated on credal frameworks, leaving representation-level comparisons largely underexplored.

\section{Preliminaries}
\label{Sec: preliminaries}
\subsection{Problem settings}
\label{Subsec: ProblemStatement}
In a supervised $K$-class classification problem, an NN with learnable parameters $\theta$, denoted by $f_{\theta}(\cdot)$, is typically trained on i.i.d. samples $\mathcal{D}=\{\boldsymbol{x}_n, y_n\}_{n=1}^{N}\subset \mathcal{X}\times \mathcal{Y}$, where $\mathcal{X}$ is the input space and $\mathcal{Y}=\{1, ..., k, ..., K\}$ is the output space. Given a test input $\boldsymbol{x}\in\mathcal{X}$, the NN outputs a softmax probability vector $\boldsymbol{p}:=\big(p(y\!=\!1|\boldsymbol{x}),..., p(y\!=\!K|\boldsymbol{x})\big)$. However, this single conditional distribution captures only aleatoric predictive uncertainty, as it assumes precise knowledge of the underlying input–output mapping~\citep{hullermeier2021aleatoric, WangTPAMI}. In this setting, representing EU generally requires a second-order formalism that expresses uncertainty about the model’s probabilistic prediction itself. The notable formalisms include distribution- and set-based representations, which are the main focus of this work.

\subsection{Uncertainty representations}
\label{Subsec: PredictorBaselines}
\textbf{Distribution-based representations.} Bayesian neural networks (BNNs) and deep ensembles (DE) are well-known distribution-based approaches. A BNN~\citep{blundell2015weight, gal2016dropout, krueger2017Bayesian, mobiny2021dropconnect} learns a posterior distribution over parameters, $p(\theta|\mathcal{D})$, obtained by applying Bayes' rule:
\begin{equation}
	p(\theta|\mathcal{D}) = \frac{p(\mathcal{D}|\theta)\,p(\theta)}{p(\mathcal{D})},
	\label{Eq:posterior_bnn}
\end{equation}
where $p(\theta)$, $p(\mathcal{D})$, and $p(\mathcal{D}|\theta)$ denote the prior over parameters, the evidence, and the likelihood, respectively. Given a test input $\boldsymbol{x}$, a BNN theoretically marginalizes over this posterior to produce a prediction:
\begin{equation}
	p\big(\boldsymbol{p}|\boldsymbol{x}, \mathcal{D}\big) = \int_{\theta} p\big(\boldsymbol{p}|\boldsymbol{x}, \theta\big)\,p(\theta|\mathcal{D})\, d\theta.
	\label{Eq:bnn_prediction}
\end{equation}
Since the network is deterministic given $\theta$, each parameter sample yields exactly one predictive distribution, $\boldsymbol{p}=f_{\theta}(\boldsymbol{x})$; that is, $p(\boldsymbol{p}|\boldsymbol{x},\theta)$ places all its probability mass on this single $\boldsymbol{p}$. The posterior $p(\theta|\mathcal{D})$ therefore induces a probability density over the predictive distributions $\boldsymbol{p}$ themselves, rather than over the model output for a fixed $\theta$. Eq.~\eqref{Eq:bnn_prediction} can thus be interpreted as a \emph{second-order distribution}---a probability distribution over probability distributions~\citep{meier2021ensemble, WangTPAMI}---representing epistemic predictive uncertainty.

However, directly calculating~\eqref{Eq:bnn_prediction} is computationally intractable. In practice, this is approximated using Bayesian model averaging (BMA)~\citep{jospin2022hands}, i.e., $M$ stochastic forward passes through the BNN are performed:
\begin{equation}
	\boldsymbol{p}_m = f_{\theta_{m}}(\boldsymbol{x}) \text{\ for\ } m=1,...,M,
	\label{Eq:MC_bnn}
\end{equation}
where $f_{\theta_{m}}(\cdot)$ denotes an NN with parameters $\theta_{m}$ sampled from the posterior $p(\theta|\mathcal{D})$ at the $m$-th pass. The predictive distribution is then approximated by averaging the probability vectors, $\tilde{\boldsymbol{p}} =\textstyle M^{-1}\sum\nolimits_{m=1}^{M}\boldsymbol{p}_m$, and the final class prediction is given by $\operatorname*{argmax}(\tilde{\boldsymbol{p}})$. Under this approximation, this distribution-based representation can be viewed as \emph{a uniform distribution over a finite set of predictive distributions}, denoted by $\mathcal{B}:=\{\boldsymbol{p}_m\}_{m=1}^M$, where the predictive distributions—rather than the underlying BNNs—are assigned equal weight.\footnote{While BMA is theoretically weighted by the posterior over BNN parameters, equal weights are assigned to the finite sampled predictive distributions in practice for approximating, e.g., the final prediction and the mutual information. Here, we emphasize that the uniform distribution here is over these finite predictive samples, not over the BNNs themselves.}

Unlike BNNs, which explicitly infer a distribution over model parameters, DE~\citep{lakshminarayanan2017simple} marginalizes over multiple models, $\{f_{\theta_m}(\cdot)\}_{m=1}^{M}$~\citep{band2benchmarking}. At inference time, DE performs single forward passes across ensemble members to produce \emph{a finite set of predictive distributions}, whose average is used to make the final class prediction. Thus, DE has been viewed as an approximation to BMA by some studies~\citep{wilson2020bayesian, abe2022deep}. Several variants—such as batch ensembles~\citep{wen2020batchensemble}, masked ensembles~\citep{durasov2021masksembles}, and packed ensembles~\citep{laurentpacked}—have been proposed to predict $\{\boldsymbol{p}_m\}_{m=1}^M$ within a single model, achieving comparable uncertainty quantification performance with a lower computational cost. 

Following common practice, although these predictions originate from different learning algorithms, we do not distinguish between them in the subsequent evaluation and analysis and denote them as $\mathcal{B}:=\{\boldsymbol{p}_m\}_{m=1}^M$ in our study.

\textbf{Set-based representations.}
Credal sets~\citep{levi1980enterprise}, denoted as convex sets of probability distributions, have been argued to provide a more natural EU representation than single probability distributions~\citep{corani2012bayesian,hullermeier2021aleatoric}. For example, sets can better capture ignorance as a lack of knowledge~\citep{dubois2002representing}, since a single distribution typically requires additional assumptions beyond merely distinguishing plausible from implausible candidates~\citep{lohr2025credal}. 

To eliminate confounding effects arising from differences in model and learning assumptions---performance differences attributable to the base model rather than to the uncertainty representation itself—our comparative study (see \figurename~\ref{FIG: Overview}) focuses on distribution- and set-based representations derived from the same underlying neural network (NN). Although various NN approaches have been proposed for generating credal predictions—including methods based on finitely generated credal sets~\citep{caprio2024credal, chau2025credal} and predicted probability intervals~\citep{wang2024CredalEnsembles,wang2025creinns,wang2026Distill}—these approaches are typically algorithm-specific. We therefore adopt a common and computationally efficient strategy. Specifically, we transform a finite set of predictive distributions into a credal set via class-wise probability intervals~\citep{wang2025credalWrapper,lohr2025credal}.

Specifically, for the $k$-th class, the upper and lower bounds of the probability interval, denoted by $p_{U_k}$ and $p_{L_k}$, respectively, are obtained from
\begin{equation}
	p_{U_k}\!=\! \operatorname*{max}_{m=1,..,M}{p_{k,m}} \quad \text{and} \quad p_{L_k}\!=\! \operatorname*{min}_{m=1,..,M}{p_{n,k}}
	\label{Eq: constructK},
\end{equation}
where $p_{n,k}$ denotes the $k$-th probability element of each $\boldsymbol{p}_n$. Thus, these probability intervals over classes define a credal set $\mathcal{K}$ as follows~\citep{probability_interval_1994}:
\begin{equation}
	\mathcal{K} = \{\boldsymbol{p} \mid p_k \in [p_{L_k}, p_{U_k}] \ \forall k =1, 2, ..., K\}
	\label{Eq: CredalPIsWrapper}.
\end{equation}
where $\boldsymbol{p}$ denotes a probability vector, and each class probability is restricted to the given probability interval.

\subsection{Epistemic uncertainty measures}
\label{Subsec: UncertaintyMeasurements}
Quantifying uncertainty requires an appropriate measure that maps a second-order prediction for a given input to a numerical value. We next introduce distinct EU measures for the two representations ($\mathcal{B}$ and $\mathcal{K}$), respectively.

\textbf{Measures for a distribution-based representation.} 
Using Shannon entropy as a classical measure of uncertainty in classification, the EU for a practical Bayesian representation, $\mathcal{B}\!=\!\{\boldsymbol{p}_m\}_{m=1}^{M}$, is computed as follows:
\begin{equation}
\sum\limits_{k=1}^{K}-\tilde{p}_{k}\log_2 \tilde{p}_{k} -
\frac{1}{M}\!\sum\limits_{m=1}^{M}\sum\limits_{k=1}^{K}\!-p_{k,m}\!\log_2 p_{k,m}.
\label{Eq: EUBNN_Class}
\end{equation}
Here, $\tilde{p}_k$ and $p_{k,m}$ are the $k$-th element of the averaged probability vector $\tilde{\boldsymbol{p}}$ and the $m$-th probability vector $\boldsymbol{p}_m$, respectively. EU is computed using the standard decomposition of total predictive uncertainty into aleatoric and epistemic parts and can be interpreted as an approximate Mutual Information (MI)~\citep{hullermeier2021aleatoric}.

Alternative to the entropy-based measure in~\eqref{Eq: EUBNN_Class}, a Label-Wise Variance (LWV) has recently been proposed~\citep{salelabel}. Given $\mathcal{B}\!=\!\{\boldsymbol{p}_m\}_{m=1}^{M}$, EU is quantified by
\begin{equation}
\sum\limits_{k=1}^{K}\tilde{p}_k(1 - \tilde{p}_k) - 
\sum\limits_{k=1}^{K}\!\frac{1}{M}\!\sum\limits_{m=1}^{M}p_{k,m}(1 - p_{k,m})
\label{Eq: EUBNN_Variance}.
\end{equation}
Here, the quantified EU is regarded as an approximation of the expected reduction in squared-error loss, analogous to mutual information, which quantifies the expected reduction in log-loss~\citep{salelabel}.

In addition, a Wasserstein Distance (WD) measure, inspired by optimal transport theory, has been proposed to quantify EU as follows~\citep{sale2024secondorder}:
\begin{equation}
    \frac{1}{2}\operatorname*{minimize}\limits_{\boldsymbol{p}^*}\sum\nolimits_{m=1}^{M}{\parallel \boldsymbol{p}_m -\boldsymbol{p}^*\parallel}_{1} 
	\label{Eq: EUBNN_WD},
\end{equation}
where $\boldsymbol{p}^* \in \Delta^{K-1}$ is the decision probability vector on the full probability simplex $\Delta^{K-1}$. Under this context, the quantified EU corresponds to the minimal Wasserstein distance between the approximated second-order prediction $\mathcal{B}$ and any possible distribution on the probability simplex of $K$ classes. In the binary case, the optimization in~\eqref{Eq: EUBNN_WD} simplifies and admits the following closed-form solution:
\begin{equation}
\sum\nolimits_{m=1}^{M}{\mid {p}_m - \operatorname*{median}(p_1,..., p_M)\mid}
\label{Eq: EUBNN_WDBinary}.
\end{equation}

\textbf{Measures for a credal representation.} To quantify EU of a credal set $\mathcal{K}$, a widely used measure is the Shannon entropy difference ($H_{\text{diff}}$)~\citep{abellan2006disaggregated}, defined as the difference between the upper and lower entropy:
\begin{equation}
\operatorname{maximize}\limits_{\boldsymbol{p} \in \mathcal{K}}H(\boldsymbol{p}) - \operatorname{minimize}\nolimits_{\boldsymbol{p} \in \mathcal{K}}H(\boldsymbol{p})
\label{Eq: EUCredal_Entropy}.
\end{equation}
Here, $H(\boldsymbol{p})$ is the classical entropy of a single probability vector. Thus, solving the maximization problem in~\eqref{Eq: EUCredal_Entropy} amounts to finding the maximum entropy over $\mathcal{K}$, that is,
\begin{equation}
	\begin{aligned}
		&\operatorname{maximize}\sum\nolimits_{k=1}^{K}-p_k\log_2p_k \quad  \text{s.t.} \\
		p_k \in&[p_{L_k}, p_{U_k}] \text{\ for\ } k=1,..K \text{\ and\ } \textstyle\sum\nolimits_{k=1}^Kp_k=1
		\label{Eq: MaximalEntropy}.
	\end{aligned}
\end{equation}
Similarly, solving the minimization problem in~\eqref{Eq: EUCredal_Entropy} requires replacing $\operatorname{maximize}$ by $\operatorname{miximize}$. In the binary case, the credal set reduces to an interval $[p_L, p_U]$, and these optimization problems admit analytical closed-form solutions, which simplify the computation of $H_{\text{diff}}$, as follows:
\begin{equation}
\begin{aligned}
&H(0.5) \!-\! \min\!\big(\!H(p_L), H(p_U\!)\big) \ \text{\emph{if}} \ 0.5 \!\in\! [p_L, p_U]  \\
&\max\!\big(\!H(p_L), H(p_U)\!\big) \!-\! \min\!\big(\!H(p_L), H(p_U)\!\big) \ \text{\emph{else}} 
\end{aligned}
\label{Eq: MaximalEntropyBinary}.
\end{equation}

An alternative measure to quantify EU of a credal set is the Generalized Hartley (GH) measure~\citep{abellan2000non}, which corresponds to the expected Hartley measure~\citep{hartley1928transmission} taken over all subsets $\mathcal{Q}$ of the output space:
\begin{equation}
\sum\limits_{\mathcal{Q} \subseteq \mathcal{Y}}\text{m}_{\mathcal{K}}(\mathcal{Q})\log_2 (|\mathcal{Q}|),
\label{Eq: EpiUncertaintyCredal}
\end{equation}
where $\text{m}_{\mathcal{K}}$ denotes the mass function induced by $\mathcal{K}$ and $|\mathcal{Q}|$ is the cardinality of $\mathcal{Q}$. The quantity $\text{m}_{\mathcal{K}}(\mathcal{Q})$ is computed from the Möbius inverse of the capacity function $\nu_{\mathcal{K}}$~\citep{hullermeier2021aleatoric}:
\begin{equation}
	\text{m}_{\mathcal{K}}(\mathcal{Q}) = \sum\limits_{\mathcal{A} \subseteq \mathcal{Q}} (-1)^{|\mathcal{Q} \backslash \mathcal{A}|} \nu_{\mathcal{K}}(\mathcal{A}),
	\label{Eq: MöbiusInverse}
\end{equation}
with $\mathcal{Q} \backslash \mathcal{A} = \{k \mid k \in \mathcal{Q}, \ k \notin\mathcal{A}\}$, and $\nu_{\mathcal{K}}(\mathcal{A})$ denoting the lower probability of $\mathcal{A} \subseteq \mathcal{Q}$. For a credal set $\mathcal{K}$ defined by probability intervals in~\eqref{Eq: CredalPIsWrapper}, $\nu_{\mathcal{K}}(\mathcal{A})$ can be computed directly as follows~\citep{probability_interval_1994}:
\begin{equation}
\begin{aligned}
\max\Big(\!\sum\nolimits_{j\in\mathcal{A}} p_{L_k}, 1-\sum\nolimits_{j\notin\mathcal{A}} p_{U_k}\Big)
\end{aligned}.
\label{Eq: lowerProbCompute}
\end{equation}
The full $\text{GH}$ calculation process~\citep{wang2024CredalEnsembles} is presented in Algorithm~\ref{alg: GH calculation} in the Appendix. 

More recently, an imprecise probability metric framework~\citep{chau2025integral} introduces the maximum mean imprecision (MMI) measure, employing the total variance distance to quantify credal epistemic uncertainty. In classification, the MMI is given as 
\begin{equation}
\sup\limits_{\mathcal{A} \subseteq \mathcal{Y}}\overline{\mathit{P}}(\mathcal{A})-\underline{\mathit{P}}(\mathcal{A}),
\label{Eq: MMI}
\end{equation}
where $\underline{\mathit{P}}(\mathcal{A})$ and $\overline{\mathit{P}}(\mathcal{A})$ denote the lower and upper probabilities of a subset $\mathcal{A}$, respectively. The lower probability $\underline{\mathit{P}}(\mathcal{A})$ can be computed from~\eqref{Eq: lowerProbCompute} for a credal set $\mathcal{K}$ in~\eqref{Eq: CredalPIsWrapper}, while $\overline{\mathit{P}}(\mathcal{A})$ is the conjugate of $\overline{\mathit{P}}(\mathcal{A})$, defined as follows:
\begin{equation}
    \overline{\mathit{P}}(\mathcal{A})=1-\underline{\mathit{P}}(\mathcal{A}^c),
	\label{Eq: UpperProbability}
\end{equation}
where $\mathcal{A}^c$ is the complement of $\mathcal{A}$ on the output space $\mathcal{Y}$. For the binary case, the MMI in~\eqref{Eq: MMI} reduces to the interval length $p_U - p_L$, which coincides with the GH measure.
\subsection{Downstream evaluation tasks}
\label{Subsec: EvaluationTasks}
Since ground-truth epistemic uncertainty (EU) is unavailable, the quality of EU quantification is generally assessed through practical downstream tasks. Following common practice, our comparative study considers two widely used benchmarks: \emph{selective prediction}~\citep{hullermeier2022quantification,chau2025integral} and \emph{out-of-distribution (OOD) detection}~\citep{wang2024CredalEnsembles, lohr2025credal}.

\textbf{Selective prediction.}
The rationale for using selective prediction to assess EU quantification quality is that an EU-aware NN is expected to assign higher EU values to misclassified samples than to correctly classified ones. In practical batch processing, instances with high EU estimates are abstained from and referred to an expert to reduce the risk of misclassification. See Algorithm~\ref{alg: SelectivePrediction} for details.

Under this setting, an accuracy–rejection curve (ARC) is used to characterize the relationship between prediction accuracy on retained samples and the rejection rate~\citep{huhn2008fr3, hullermeier2022quantification}. Reliable uncertainty estimates produce a monotonically increasing ARC, whereas random rejection results in a flat curve~\citep{hullermeier2022quantification}. In addition, the area under the accuracy–rejection curve (AUARC) provides a scalar summary metric~\citep{jaegercall}, where larger AUARC values indicate stronger selective prediction performance.

\textbf{OOD detection.} As a practical benchmark for evaluating EU quantification quality, stronger OOD detection performance suggests that the estimated uncertainty is more informative~\citep{lohr2025credal,wang2026learning}. The intuition is that accurate EU estimation helps avoid misclassifying ambiguous in-distribution (ID) samples as OOD instances. Such ambiguity does not stem from regions of higher EU within the ID distribution, so a valid EU estimate should distinguish these cases~\citep{mukhoti2023deep}.

In this setting, as summarized in Algorithm~\ref{alg: ood-algorithm}, OOD detection is formulated as a binary classification problem where ID and OOD samples are assigned labels 0 and 1, respectively. The model’s EU estimate is used as the prediction score, and performance is evaluated using the area under the receiver operating characteristic curve (AUROC). A higher AUROC value indicates a better performance.
\begin{table*}[t]
	\caption{Average AUARC on the selective prediction task, computed per dataset and averaged across six predictive model architectures. We additionally report net wins, obtained from pairwise comparisons both within the same uncertainty-representation category (intra-representation) and across different categories (inter-representation). For both AUARC and net wins, higher values indicate better performance. In each column, the best-performing method is highlighted in \textbf{\textcolor{red}{bold red}} and the second-best in \textbf{\textcolor{blue}{bold blue}}. Detailed AUARC score as well as the intra- and inter-representation comparisons for each underlying predictive model are provided in Tables~\ref{Table: AUROCscores}, ~\ref{Table: NetWinsIntraSP}, and~\ref{Table: NetWinsInterSP}, respectively.}
	\label{Table: SPresultSummary}
	\centering
	\small
	\setlength\tabcolsep{4.5pt}
	\begin{tabular}{@{}lccc|ccc|ccc|ccc@{}}
		\toprule
		\multirow{2}{*}{}   & \multicolumn{3}{c|}{In-distribution Camelyon17}                                                & \multicolumn{3}{c|}{Distribution-shift Camelyon17}                                             & \multicolumn{3}{c|}{In-distribution SeaShip}                                                   & \multicolumn{3}{c}{In-distribution CIFAR10}                                                    \\ \cmidrule(l){2-13} 
		& Average scores                                      & Intra                  & Inter & Average scores                                      & Intra                  & Inter & Average scores                                      & Intra                  & Inter & Average scores                                      & Intra                  & Inter \\ \midrule
		MI                  & \multicolumn{1}{c|}{0.9602\scriptsize{$\pm$0.0568}} & \multicolumn{1}{c|}{-11}  & -18     & \multicolumn{1}{c|}{0.9637\scriptsize{$\pm$0.0176}} & \multicolumn{1}{c|}{-12}  & -19     & \multicolumn{1}{c|}{0.9887\scriptsize{$\pm$0.0156}} & \multicolumn{1}{c|}{-12}  & -18     & \multicolumn{1}{c|}{0.9807\scriptsize{$\pm$0.0050}} & \multicolumn{1}{c|}{-12}  & -18    \\
		LWV                 & \multicolumn{1}{c|}{0.9625\scriptsize{$\pm$0.0532}} & \multicolumn{1}{c|}{\textcolor{blue}{\textbf{0}}} & -5     & \multicolumn{1}{c|}{0.9669\scriptsize{$\pm$0.0131}} & \multicolumn{1}{c|}{\textcolor{blue}{\textbf{0}}} &-6     & \multicolumn{1}{c|}{0.9899\scriptsize{$\pm$0.0151}} & \multicolumn{1}{c|}{\textcolor{blue}{\textbf{0}}} & -4     & \multicolumn{1}{c|}{0.9815\scriptsize{$\pm$0.0049}} & \multicolumn{1}{c|}{\textcolor{blue}{\textbf{0}}} & -6     \\
		WD                  & \multicolumn{1}{c|}{0.9633\scriptsize{$\pm$0.0530}} & \multicolumn{1}{c|}{\textcolor{red}{\textbf{11}}} & \textcolor{red}{\textbf{21}}    & \multicolumn{1}{c|}{0.9683\scriptsize{$\pm$0.0125}} & \multicolumn{1}{c|}{\textcolor{red}{\textbf{12}}} & \textcolor{red}{\textbf{30}}     & \multicolumn{1}{c|}{0.9904\scriptsize{$\pm$0.0144}} & \multicolumn{1}{c|}{\textcolor{red}{\textbf{12}}} & \textcolor{red}{\textbf{25}}     & \multicolumn{1}{c|}{0.9823\scriptsize{$\pm$0.0046}} & \multicolumn{1}{c|}{\textcolor{red}{\textbf{12}}} & \textcolor{red}{\textbf{28}}    \\ \cmidrule(l){3-3}\cmidrule(lr){6-6}\cmidrule(lr){9-9}\cmidrule(lr){12-12}
		GH                  & \multicolumn{1}{c|}{0.9633\scriptsize{$\pm$0.0523}} & \multicolumn{1}{c|}{\textcolor{red}{\textbf{6}}} & \textcolor{blue}{\textbf{15}}    & \multicolumn{1}{c|}{0.9680\scriptsize{$\pm$0.0126}} & \multicolumn{1}{c|}{\textcolor{red}{\textbf{10}}} & \textcolor{blue}{\textbf{12}}    & \multicolumn{1}{c|}{0.9903\scriptsize{$\pm$0.0144}} & \multicolumn{1}{c|}{\textcolor{red}{\textbf{12}}} & \textcolor{blue}{\textbf{19}}     & \multicolumn{1}{c|}{0.9822\scriptsize{$\pm$0.0046}} & \multicolumn{1}{c|}{\textcolor{red}{\textbf{12}}} & \textcolor{blue}{\textbf{20}}    \\
		$H_{\text{diff}}$ & \multicolumn{1}{c|}{0.9594\scriptsize{$\pm$0.0517}} & \multicolumn{1}{c|}{-12}  & -28     & \multicolumn{1}{c|}{0.9613\scriptsize{$\pm$0.0164}} & \multicolumn{1}{c|}{-12}  & -29      & \multicolumn{1}{c|}{0.9848\scriptsize{$\pm$0.0237}} & \multicolumn{1}{c|}{-12}  & -30      & \multicolumn{1}{c|}{0.9764\scriptsize{$\pm$0.0068}} & \multicolumn{1}{c|}{-12}  & -30      \\
		MMI                 & \multicolumn{1}{c|}{0.9633\scriptsize{$\pm$0.0523}} & \multicolumn{1}{c|}{\textcolor{red}{\textbf{6}}} & \textcolor{blue}{\textbf{15}}    & \multicolumn{1}{c|}{0.9680\scriptsize{$\pm$0.0126}} & \multicolumn{1}{c|}{\textcolor{blue}{\textbf{2}}} & \textcolor{blue}{\textbf{12}}     & \multicolumn{1}{c|}{0.9901\scriptsize{$\pm$0.0149}} & \multicolumn{1}{c|}{\textcolor{blue}{\textbf{0}}} & 8     & \multicolumn{1}{c|}{0.9819\scriptsize{$\pm$0.0048}} & \multicolumn{1}{c|}{\textcolor{blue}{\textbf{0}}}& 6     \\ \bottomrule
	\end{tabular}
\end{table*}
\section{Experiments\protect\footnotemark}
\footnotetext{Code is  at:~\url{https://github.com/Kaizheng-WANG/set-vs-distribution-epistemic-representation}.}
\label{Sec: Experiment}
\textbf{Predictive models.} As illustrated in \figurename~\ref{FIG: Overview}, our comparative study constructs a \emph{finite set of predictive distributions} from a \emph{common second-order predictor}. The goal is not to contrast distinct inference paradigms, but to ensure robustness and fairness of the analysis across representative implementations. We consider the following model families: \textbf{i)}~Stochastic variational inference (SVI)~\citep{blundell2015weight,graves2011practical}, a classical BNN method that approximates the parameter posterior with a Gaussian distribution. \textbf{ii)}~Monte Carlo Dropout (MCDO)~\citep{gal2016dropout}, which estimates the posterior via stochastic forward passes with dropout enabled. \textbf{iii)}~Deep ensembles (DE)~\citep{lakshminarayanan2017simple}, which approximate Bayesian inference by marginalizing predictions from independently trained models. In addition, we include three recent, computationally efficient DE variants: \textbf{iv)}~Batch ensembles (BatchEns)~\citep{wen2020batchensemble}, which factorize each weight matrix into a shared component and a rank-one, member-specific term; \textbf{v)}~Masked ensembles (MaskEns)~\citep{durasov2021masksembles}, which use fixed binary masks to control correlations between ensemble members; and \textbf{vi)}~Packed ensembles (PackEns)~\citep{laurentpacked}, which exploit grouped convolutions to parallelize ensemble members within a shared backbone.

Specifically, for each experiment setting, DE is constructed by training $5$ individual neural networks ($M\!=\!5$) with different random seeds, following standard practice. BatchEns, MaskEns, and PackEns use the default ensemble size of $M\!=\!4$. For SVI and MCDO, we perform inference with $10$ forward passes ($M\!=\!10$). All underlying models are trained independently for $10$ runs per setting.

\textbf{Datasets and benchmarks.} We evaluate our approach on CIFAR10~\citep{cifar10} and two real-world settings: \textbf{i)}~a medical diagnosis task using whole-slide images from the Camelyon17 dataset~\citep{bandi2018detection}; \textbf{ii)}~a ship classification task reflecting realistic visual variability. The downstream tasks include selective prediction and OOD detection, see Section~\ref{Subsec: EvaluationTasks}. 

For \textbf{CIFAR10}, all models are trained on the standard training split, with selective prediction evaluated on the test split. For out-of-distribution (OOD) detection, CIFAR10 is treated as the in-distribution (ID) dataset, while SVHN and FMNIST serve as OOD benchmarks. For \textbf{medical image classification}, Camelyon17 comprises histopathology images collected from five medical centers in the Netherlands and scanned using three devices, naturally inducing realistic distribution shifts. We treat centers 0, 1, and 3 (3DHistech scanners) as ID data, and centers 2 and 4 (Philips and Hamamatsu scanners) as the distribution-shift set. Dataset statistics are reported in Table~\ref{Table: Datasplit}. Selective prediction is evaluated on both the ID and distribution-shift test splits. For \textbf{ship classification}, models are trained on the SeaShip training set and evaluated on the corresponding test split. For OOD detection, we use SeaShip-C (corruption-based shifts) and SeaShip-O, which contains ship images sourced from external datasets (SMD and SSAVE). These classification datasets are derived from ship detection benchmarks; preprocessing details are provided in Appendix~\ref{App: Ship}.

Across all datasets, predictive models are trained under a shared protocol. Detailed configurations and experimental settings are deferred to Appendix~\ref{App: DetailsExp}.
\begin{table*}[t]
	\caption{Average AUROC on the OOD detection task, computed per dataset pair and averaged across six predictive model architectures. We additionally report net wins, obtained from pairwise comparisons both within the same uncertainty-representation category (intra-representation) and across different categories (inter-representation). For both AUROC and net wins, higher values indicate better performance. In each column, the best-performing method is highlighted in \textbf{\textcolor{red}{bold red}} and the second-best in \textbf{\textcolor{blue}{bold blue}}. Detailed AUROC score as well as the intra- and inter-representation net-win comparisons for each underlying predictive model are provided in Tables~\ref{Table: AUROCscores}, ~\ref{Table: NetWinsIntraOOD}, and~\ref{Table: NetWinsInterOOD}, respectively.}
	\label{Table: OODresultSummary}
	\centering
	\small
	\setlength\tabcolsep{4.5pt}
	\begin{tabular}{@{}lccc|ccc|ccc|ccc@{}}
		\toprule
		\multirow{2}{*}{}   & \multicolumn{3}{c|}{SeaShip v.s. SeaShip-O}                                                                      & \multicolumn{3}{c|}{SeaShip v.s. SeaShip-C}                                                                      & \multicolumn{3}{c|}{CIFAR10 v.s. SVHN}                                                                           & \multicolumn{3}{c}{CIFAR10 v.s. FMNIST}                                                                          \\ \cmidrule(l){2-13} 
		& Average scores                                                        & Intra                  & Inter & Average scores                                                        & Intra                  & Inter & Average scores                                                        & Intra                  & Inter & Average scores                                                        & Intra                  & Inter \\ \midrule 
		MI                  & \multicolumn{1}{c|}{0.8694\scriptsize{$\pm$0.0840}} & \multicolumn{1}{c|}{\textcolor{blue}{\textbf{-3}}} & -6     & \multicolumn{1}{c|}{0.8250\scriptsize{$\pm$0.0859}} & \multicolumn{1}{c|}{\textcolor{blue}{\textbf{-5}}} & -11     & \multicolumn{1}{c|}{0.7657\scriptsize{$\pm$0.0301}} & \multicolumn{1}{c|}{-8} & -25    & \multicolumn{1}{c|}{0.8639\scriptsize{$\pm$0.0276}} & \multicolumn{1}{c|}{\textcolor{blue}{\textbf{0}}} & -11     \\
		LWV                 & \multicolumn{1}{c|}{0.8655\scriptsize{$\pm$0.0797}} & \multicolumn{1}{c|}{-9}  & -15     & \multicolumn{1}{c|}{0.8300\scriptsize{$\pm$0.0776}} & \multicolumn{1}{c|}{-7} & -13     & \multicolumn{1}{c|}{0.7694\scriptsize{$\pm$0.0295}} & \multicolumn{1}{c|}{\textcolor{blue}{\textbf{-4}}} & -21 & \multicolumn{1}{c|}{0.8466\scriptsize{$\pm$0.0256}} & \multicolumn{1}{c|}{-12}  & -30     \\
		WD                  & \multicolumn{1}{c|}{0.8822\scriptsize{$\pm$0.0739}} & \multicolumn{1}{c|}{\textcolor{red}{\textbf{12}}} & \textcolor{blue}{\textbf{19}}     & \multicolumn{1}{c|}{0.8512\scriptsize{$\pm$0.0720}} & \multicolumn{1}{c|}{\textcolor{red}{\textbf{12}}} & \textcolor{blue}{\textbf{18}}       & \multicolumn{1}{c|}{0.8174\scriptsize{$\pm$0.0227}} & \multicolumn{1}{c|}{\textcolor{red}{\textbf{12}}} & \textcolor{blue}{\textbf{18}}       & \multicolumn{1}{c|}{0.8731\scriptsize{$\pm$0.0232}} & \multicolumn{1}{c|}{\textcolor{red}{\textbf{12}}} &10    \\ \cmidrule(lr){3-3}\cmidrule(lr){6-6}\cmidrule(lr){9-9}\cmidrule(lr){12-12}
		GH                  & \multicolumn{1}{c|}{0.8856\scriptsize{$\pm$0.0720}} & \multicolumn{1}{c|}{\textcolor{red}{\textbf{12}}} & \textcolor{red}{\textbf{29}}    & \multicolumn{1}{c|}{0.8561\scriptsize{$\pm$0.0703}} & \multicolumn{1}{c|}{\textcolor{red}{\textbf{12}}} & \textcolor{red}{\textbf{30}}     & \multicolumn{1}{c|}{0.8303\scriptsize{$\pm$0.0197}} & \multicolumn{1}{c|}{\textcolor{red}{\textbf{12}}} & \textcolor{red}{\textbf{30}}    & \multicolumn{1}{c|}{0.8801\scriptsize{$\pm$0.0226}} & \multicolumn{1}{c|}{\textcolor{red}{\textbf{9}}} & \textcolor{red}{\textbf{27}}    \\
		$H_{\mathrm{diff}}$ & \multicolumn{1}{c|}{0.8516\scriptsize{$\pm$0.0893}} & \multicolumn{1}{c|}{-12}  & -30      & \multicolumn{1}{c|}{0.7929\scriptsize{$\pm$0.0912}} & \multicolumn{1}{c|}{-12}  & -30     & \multicolumn{1}{c|}{0.7797\scriptsize{$\pm$0.0343}} & \multicolumn{1}{c|}{-12}  & -8    & \multicolumn{1}{c|}{0.8744\scriptsize{$\pm$0.0266}} & \multicolumn{1}{c|}{\textcolor{blue}{\textbf{3}}} & \textcolor{blue}{\textbf{17}}    \\
		MMI                 & \multicolumn{1}{c|}{0.8768\scriptsize{$\pm$0.0751}} & \multicolumn{1}{c|}{\textcolor{blue}{\textbf{0}}} & 3     & \multicolumn{1}{c|}{0.8460\scriptsize{$\pm$0.0729}} & \multicolumn{1}{c|}{\textcolor{blue}{\textbf{0}}} &6     & \multicolumn{1}{c|}{0.8038\scriptsize{$\pm$0.0235}} & \multicolumn{1}{c|}{\textcolor{blue}{\textbf{0}}} & 6    & \multicolumn{1}{c|}{0.8630\scriptsize{$\pm$0.0237}} & \multicolumn{1}{c|}{-12}  & -13    \\ \bottomrule
	\end{tabular}
\end{table*}
\section{Comparative analysis}
\label{Sec: ResultsAnalysis}
\subsection{Evaluation criteria and results}
\textbf{Evaluation criteria.} For each downstream task and dataset, we evaluate all second-order predictive models (DE, SVI, MCDO, BatchEns, MaskEns, and PackEns). For distribution-based representations, uncertainty is quantified using Mutual Information (MI) in~\eqref{Eq: EUBNN_Class}, Label-wise Variance (LWV) in~\eqref{Eq: EUBNN_WD}, and Wasserstein Distance (WD) in~\eqref{Eq: EUBNN_WDBinary} for uncertainty quantification. For credal representations, we instead consider the entropy difference ($H_{\text{diff}}$) in~\eqref{Eq: EUCredal_Entropy}, the Generalized Hartley (GH) measure in~\eqref{Eq: EpiUncertaintyCredal}, and the Maximum Mean Imprecision (MMI) in~\eqref{Eq: MMI}. Additional details are provided in Section~\ref{Subsec: UncertaintyMeasurements}.

Beyond reporting average performance scores (AUARC for selective prediction and AUROC for OOD detection, respectively), we perform pairwise one-sided Wilcoxon signed-rank tests at the 5$\%$ significance level. For each ordered pair of uncertainty measures $(m_i, m_j)$ with $i \neq j$, the null hypothesis ($H_0$) assumes no systematic performance difference between $m_i$ and $m_j$, while the alternative hypothesis ($H_1$) asserts that $m_i$ yields stochastically larger performance scores than $m_j$. The tests are conducted over 10 independent runs. Results are considered statistically significant when $p < 0.05$, in which case $m_i$ is deemed to outperform $m_j$.

Based on the statistical tests, we construct a ranking scheme to provide an interpretable quantitative summary. For each predictive model and uncertainty measure, we record the number of significant wins and losses across pairwise comparisons. Each significant win contributes $+1$, each significant loss contributes $-1$, and non-significant outcomes contribute $0$. The resulting net score, defined as $\text{wins}(m) - \text{losses}(m)$, captures the relative dominance of a measure. Global rankings, including both intra- and inter-representation comparisons, are obtained by aggregating net scores across predictive models.

\begin{figure*}[t]
\centering
\includegraphics[width=1\linewidth]{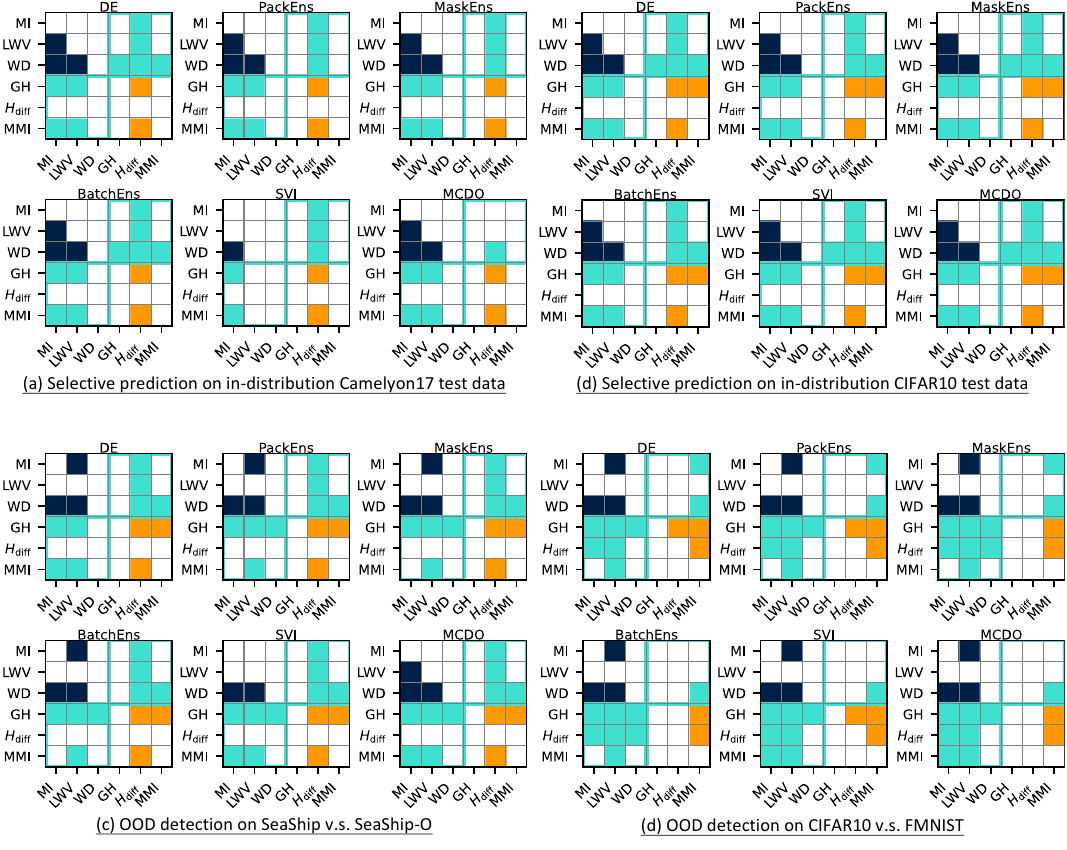}
\caption{Statistical significance plots on different selective prediction (a, b) and OOD detection (c, d) benchmarks across different underlying predictive models. A cell is shaded if the measure in the $i$-th row is statistically significantly better than that in the $j$-th column according to a pairwise one-sided Wilcoxon signed-rank test at the 5\% significance level. Intra-representation comparisons are shown in blue (distribution-based measures) and orange (credal-based measures), while inter-representation comparisons are shown in green.}
\label{FIG: SPOODresultsParital}
\end{figure*}

\textbf{Results.} For the selective prediction task, \tablename~\ref{Table: SPresultSummary} reports the average AUARC across six underlying predictive models and 10 runs. For the OOD task, \tablename~\ref{Table: OODresultSummary} reports the average AUROC under the same setting. Both tables also summarize the net wins across the six underlying predictive models for intra- and inter-representation comparisons, based on one-sided paired Wilcoxon tests at the 5\% significance level (partially shown in \figurename~\ref{FIG: SPOODresultsParital} and fully presented in Figures~\ref{FIG: SPresults} and~\ref{FIG: OODresults}).

The accuracy-rejection curves for selective prediction are presented in Figures~\ref{FIG: MedicalIDARC}, \ref{FIG: MedicalOODARC}, \ref{FIG: ShipARC}, and \ref{FIG: CIFARARC}, while Receiver Operating Characteristic (ROC) curves for OOD detection are presented in Figures~\ref{FIG: ROCShipO}, \ref{FIG: ROCShipC}, \ref{FIG: ROCFMNIST}, and \ref{FIG: ROCSVHN}.

\subsection{Summary and analysis}
\label{subsec: summary}
\textbf{(1) The relative merits of an uncertainty representation cannot be assessed independently of the associated uncertainty measures.} 

\tablename~\ref{Table: SPresultSummary} and \tablename~\ref{Table: OODresultSummary} show that neither representation consistently dominates across benchmarks. Performance varies systematically with the uncertainty measure. For instance, the credal representation paired with the GH measure attains the strongest results on OOD detection, whereas the distribution-based representation combined with the WD measure ranks highest on selective prediction under the statistical tests.

While our inclusion of multiple second-order predictive models and datasets is intended to improve the robustness of the analysis rather than compare underlying models, \figurename~\ref{FIG: SPOODresultsParital} shows that uncertainty-aware performance remains sensitive to predictive model choices and dataset. Even when fixing the representation, uncertainty measure, downstream task, and dataset, results vary across different predictive models. Similarly, holding the representation, uncertainty measure, predictive model, and task constant while changing datasets yields different outcomes under the one-sided paired Wilcoxon tests.

Taken together, these findings highlight that empirical comparisons of uncertainty representations must be interpreted conditionally. Claims of effectiveness should therefore be qualified with explicit reference to the uncertainty measure, benchmark, predictive model, and dataset.

\textbf{(2) OOD detection more readily reveals differences between the two uncertainty representations than selective prediction.} 

Although one-sided paired Wilcoxon tests indicate measurable differences on selective prediction benchmarks, the gaps in average AUARC remain small (Table~\ref{Table: SPresultSummary}). This behavior follows naturally from the evaluation protocol. As discussed in Section~\ref{Subsec: EvaluationTasks}, selective prediction combines instance rejection with accuracy on the retained samples. Distinct uncertainty estimates can therefore yield similar rejection sets, leading to nearly identical performance. Even when rejection patterns differ, predictive accuracy often changes only marginally because all measures operate on the same underlying model. Moreover, baseline accuracy without rejection is already high (Figure~\ref{FIG: ShipARC}), which further compresses observable gains.

By contrast, OOD detection constitutes a more sensitive regime for differentiating uncertainty representations, as detection performance directly depends on how uncertainty responds to distributional shifts, despite known limitations of EU-based OOD methods~\citep{li2025position}.

These observations suggest that validating new uncertainty representations or measures across multiple downstream tasks is important for establishing robust empirical claims.

\textbf{(3) Reliable uncertainty quantification depends jointly on the representation and the uncertainty measure.}

For distribution-based representations, \tablename~\ref{Table: SPresultSummary} and \tablename~\ref{Table: OODresultSummary} show that WD consistently outperforms MI and LWV across both selective prediction and OOD detection benchmarks. A plausible explanation is that WD more directly captures epistemic predictive uncertainty by quantifying the geometric dispersion of the second-order distribution around its barycenter. This captures variability induced by individual predictive distributions. In contrast, MI relies on entropy-based uncertainty decomposition and is influenced by the global shape of the predictive distribution, while LWV measures label-wise variability without accounting for the geometry of the probability simplex. Moreover, prior analysis suggests that~\citet{sale2024secondorder} WD better satisfies key theoretical desiderata than MI and LWV.

For credal representations, GH consistently achieves the strongest performance. This likely stems from its more expressive characterization of set-valued uncertainty, as GH evaluates the structure of the credal set rather than a single scalar summary. By comparison, the entropy difference $H_{\text{diff}}$, despite its popularity, performs poorly in most settings. One limitation is that it depends solely on the width of the Shannon entropy interval. For example, in a 2D probability simplex, any credal set that includes the center and one vertex results in the same EU, regardless of the shape of the set itself. This behavior is not attributable to numerical optimization artifacts, as closed-form solutions exist in the binary case. We note, however, that GH becomes computationally impractical for a large label space in $K$ classification, requiring calculations over $2^K$ subsets. For theoretical discussions on GH and $H_{\text{diff}}$, see~\citet{hullermeier2022quantification}. This also leads to the recent development of MMI and its linear time upper bound, as an alternative approach to GH~\citep{chau2025integral}. 

Overall, these findings emphasize that improving uncertainty quantification requires not only principled representations but also carefully designed and computationally efficient uncertainty measures.

\paragraph{Further Experiments.} To further examine generalizability, we conduct reciprocal experiments in which SVHN and FMNIST are each treated as the ID dataset in turn, with the remaining dataset and CIFAR10 serving as OOD counterparts. The same six predictive models are trained on FMNIST and SVHN, then evaluated on both selective prediction and OOD detection under the same protocol. Detailed results are reported in Appendix~\ref{App: furtherExps}; they consistently support and further strengthen the generalizability of our three main findings discussed in Section~\ref{subsec: summary}.

\section{Conclusion}
\label{Sec: DiscussConclude}
This paper present a systematic comparison of distribution-based and credal uncertainty representations for classification. To isolate representational effects, both formalisms were derived from an identical finite set of predictive distributions produced by shared predictive models. The study covers $6$ uncertainty measures, $14$ selective prediction and OOD detection benchmarks, $6$ predictive model families, and $10$ independent runs per configuration.

Three main observations emerge from the empirical analysis. First, neither representation demonstrates uniform superiority; performance depends critically on the interaction between representation and uncertainty measure. Second, OOD detection more readily reveals representational differences than selective prediction, highlighting task-dependent sensitivity of uncertainty evaluation. Third, reliable uncertainty quantification depends jointly on representation and metric choice, as different measures induce markedly different behaviors even under fixed predictive models.

These results underscore the importance of conditional interpretation in empirical studies of uncertainty. Claims regarding uncertainty representations or EU-aware predictors should therefore be grounded in clearly specified evaluation settings, including the uncertainty measure, benchmark, and dataset. Robust validation further benefits from assessing multiple downstream tasks. Finally, our findings highlight that progress in uncertainty-aware learning is driven not only by representational advances but also by the design of theoretically grounded and computationally efficient uncertainty measures.

\textbf{Limitations. }Alternative second-order representations for classification, such as Dirichlet-based models and random sets, were not included in this study. While these frameworks are conceptually related, they lack a general, representation-agnostic construction from arbitrary EU-aware predictors and do not admit straightforward translation across formalisms. This complicates controlled, like-for-like comparisons under a shared experimental protocol. Consequently, we restrict attention to representations that can be consistently derived from a common finite set of predictive distributions. For similar reasons, we focus on probability-interval credal sets induced by finite predictive samples and do not consider alternative credal constructions explored in other credal classification frameworks.

Computational considerations further constrain the scale of the empirical analysis. Extending the evaluation to a wider range of uncertainty measures, datasets, second-order predictors, and benchmark settings (e.g., active learning) remains an important direction for future work.

\begin{acknowledgements} 
We thank the anonymous reviewers for their valuable feedback. This work was supported by the Start-up Grant from Nanyang Technological University, Singapore, and by the European Union's Horizon 2020 research and innovation programme under grant agreement No 964505 (E-pi) and under the Marie Sklodowska-Curie grant agreement No 955768 (AUTOBarge). This research was also partially supported by Flanders Make, the strategic research center for the manufacturing industry.
\end{acknowledgements}

\bibliography{uai2026}

\newpage

\onecolumn
\makeatletter
\def\thanks#1{\footnotemark} 
\makeatother
\title{Set-based v.s. Distribution-based Representations of Epistemic Uncertainty:\\ A Comparative Study\\(Supplementary Material)}
\maketitle
\appendix
\renewcommand{\theequation}{A.\arabic{equation}}
\setcounter{equation}{0} 
\renewcommand{\thefigure}{A.\arabic{figure}}
\setcounter{figure}{0} 
\renewcommand{\thetable}{A.\arabic{table}}
\setcounter{table}{0} 
\section{Implementation details}
\subsection{Experimental configurations}
\label{App: DetailsExp}
\textbf{Configurations of predictive models.}
In our comparative study, we consider several second-order predictive models:
\textbf{i)}~Stochastic variational inference (SVI)~\citep{blundell2015weight,graves2011practical}, a classical BNN approach that approximates the parameter posterior with a Gaussian distribution.
\textbf{ii)}~Monte Carlo Dropout (MCDO)~\citep{gal2016dropout}, which estimates the posterior through stochastic forward passes with dropout enabled at inference time.
\textbf{iii)}~Deep ensembles (DE)~\citep{lakshminarayanan2017simple}, which approximate Bayesian inference by marginalizing predictions from independently trained models.

In addition, we include three recent computationally efficient variants of DE:
\textbf{iv)}~Batch ensembles (BatchEns)~\citep{wen2020batchensemble}, which factorize each weight matrix into a shared component and a rank-one, member-specific term;
\textbf{v)}~Masked ensembles (MaskEns)~\citep{durasov2021masksembles}, which use fixed binary masks to control correlations among ensemble members; and
\textbf{vi)}~Packed ensembles (PackEns)~\citep{laurentpacked}, which leverage grouped convolutions to parallelize ensemble members within a shared backbone.

All predictors use a ResNet architecture~\citep{he2016deep} as the backbone (ResNet-34 for the SeaShip and Camelyon17 datasets, and ResNet-18 for CIFAR10), with an input size of $(3, 224, 224)$. For DE, we train $M=5$ independent neural networks with different random seeds. The implementations of the ensemble members, SVI, and MCDO follow the guidelines of the repository~\url{https://github.com/DBO-DKFZ/uncertainty-benchmark}. BatchEns, MaskEns, and PackEns are implemented according to the repository~\url{https://torch-uncertainty.github.io/}, using the default ensemble size $M=4$. For SVI and MCDO, inference is performed using $M=10$ stochastic forward passes.

\textbf{Training configurations.}
The training batch size is set to 128. We use the Adam optimizer with an initial learning rate of $0.001$, which is reduced by a factor of $10$ if the validation cross-entropy loss does not improve for three consecutive epochs. Data preprocessing, augmentation, and other training procedures follow the protocol described in~\citet{mehrtens2023benchmarking}. Specifically, for the Camelyon17 and SeaShip datasets, we apply the strong augmentation mode, while for the relatively simpler CIFAR10 dataset, we use the crop mode.

Each predictive model is trained for up to $30$, $60$, and $100$ epochs on Camelyon17, SeaShip, and CIFAR10, respectively. For evaluation, we select the checkpoint with the best balanced validation accuracy. All models are trained on a single NVIDIA P100 SXM2@1.3, GHz GPU. An exception is BatchEns on Camelyon17, which requires an NVIDIA V100 SXM2@1.5,GHz GPU due to higher memory consumption.

To support reproducibility, the core implementation code for running and analyzing the experiments will be released publicly upon publication under a license that allows free use for research.

\textbf{Additional information for experiments on Camelyon17 dataset.}
In our evaluation on medical classification, we focus on a challenging real-world medical diagnosis task involving whole-slide images (WSIs). The main difficulties arise from \textbf{i)} the enormous size of WSIs combined with the limited availability of annotated data, and \textbf{ii)} distribution shifts due to differences in image acquisition across institutions and scanners, where deployment data often deviates from the training distribution~\citep{mehrtens2023benchmarking}.

\begin{figure}[!htbp]
\begin{center}
\includegraphics[width=0.65\linewidth]{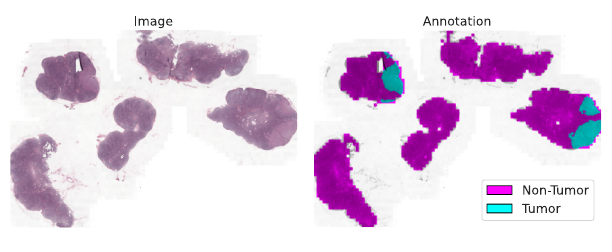}
\end{center}
\caption{An example of the whole slide image (referring to node 2 of patient 017) with ground-truth annotations.}
\label{FIG: WSIimage}
\end{figure}

For our experiments, we use the Camelyon17 dataset~\citep{bandi2018detection}, which consists of 50 breast lymph node WSIs with metastatic tissue, collected from five medical centers in the Netherlands and scanned on three different devices. An example WSI is shown in \figurename~\ref{FIG: WSIimage}. 
To simulate a strong domain shift~\citep{mehrtens2023benchmarking}, we partition the dataset as follows: centers 0, 1, and 3—scanned with 3DHistech devices—are grouped as in-distribution (ID) data, while centers 2 and 4—scanned with Philips and Hamamatsu devices—serve as domain-shifted data. Following the protocol described in~\citep{mehrtens2023benchmarking, khened2021generalized}, lesion-level tile instances are extracted from the WSIs, which are then used for network training, validation, and testing. The resulting dataset statistics are summarized in \tablename~\ref{Table: Datasplit}.
\begin{table}[!htbp]
\caption{Number of tile instances in each dataset split.}
\label{Table: Datasplit}
\centering
\small
\setlength\tabcolsep{5pt}
\begin{tabular}{@{}ccc|cc@{}}
\toprule
\multicolumn{3}{c|}{ID dataset} & \multicolumn{2}{c}{Domain-shifted test dataset} \\ \midrule
Training & Validation & Testing & Center 2          & Center 4         \\ \midrule
383406   & 110561     & 109060  & 89351             & 166607           \\ \bottomrule
\end{tabular}
\end{table}
\subsection{Algorithmic implementations}
\begin{algorithm}[!htbp]
	\begin{algorithmic}
		\renewcommand{\algorithmiccomment}[1]{\hskip0em$\#$ {#1}}
		\STATE \textbf{Input:} $[\boldsymbol{p}_L, \boldsymbol{p}_U]\!:=\! \{[p_{L_k}, p_{U_k}]\}_{k=1}^{K}$; Target space $\mathcal{Y}$
		\STATE \textbf{Output:} $\text{GH}(\mathcal{K})$ 
		\STATE Initialize: $\text{GH}(\mathcal{K})\!=\!0$
		\FORALL{$\mathcal{A} \subseteq \mathcal{Y} \ \AND \ |\mathcal{A}|\geq2$}
		\STATE Initialize: $\text{m}_{\mathcal{K}}(\mathcal{A}) = 0$
		\FORALL{$\mathcal{Q} \subseteq \mathcal{A}$}
		\STATE Compute $\nu_{\mathcal{K}}(\mathcal{Q})$ using~\eqref{Eq: lowerProbCompute}
		\STATE  $\text{m}_{\mathcal{K}}(\mathcal{A}) = \text{m}_{\mathcal{K}}(\mathcal{A}) \!+\! (-1)^{|\mathcal{A} \backslash \mathcal{Q}|}\!\cdot\!\nu_{\mathcal{K}}(\mathcal{Q})$   using~\eqref{Eq: MöbiusInverse}
		\ENDFOR
		\STATE  $\text{GH}(\mathcal{K}) \!=\! \text{GH}(\mathcal{K}) \!+\! \text{m}_{\mathcal{K}}(\mathcal{A})\!\cdot\!\log_2(|\mathcal{A}|)$ using~\eqref{Eq: EpiUncertaintyCredal}
		\ENDFOR
		\caption{$\text{GH}$ Calculation Procedure}
		\label{alg: GH calculation}
	\end{algorithmic}
\end{algorithm}

\begin{algorithm}[!hbtp]
\begin{algorithmic}
\STATE\textbf{Input:} Test dataset $\mathcal{D}_{\text{test}}$; rejection rate $\beta$; 
second-order predictor $f(\cdot)$
\STATE \textbf{1.} Obtain second-order representations from $f(\cdot)$ on $\forall \boldsymbol{x}_n \in \mathcal{D}_{\text{test}}$
\STATE $\mathcal{B}_n\leftarrow f(\boldsymbol{x}_n)$ (practical Bayesian) or $\mathcal{K}_n\leftarrow\mathcal{B}_n$ using~\eqref{Eq: constructK} (credal)
\STATE \textbf{2.} Quantify epistemic predictive uncertainty given each sample, denoted by $u_{\text{EU},n}$
\STATE $u_{\text{EU},n} \forall n $ using~\eqref{Eq: EUBNN_Class}/\eqref{Eq: EUBNN_Variance}/\eqref{Eq: EUBNN_WD} for $\mathcal{B}_n$ or~\eqref{Eq: EUCredal_Entropy}/\eqref{Eq: EpiUncertaintyCredal}/\eqref{Eq: MMI} for $\mathcal{K}_n$
\STATE \textbf{3.} Sort uncertainty estimates in ascending order
\STATE let $\boldsymbol{\pi}=\{1, 2, ..., N_t\}$ with $N_t = |\mathcal{D}_{\text{test}}|$ so that $u_{\text{EU},\pi(1)}\leq u_{\text{EU},\pi(2)} \leq ... \leq u_{\text{EU},\pi(N_t)}$
\STATE \textbf{4.} Select top $\lfloor (1-\beta)N_t \rfloor$ certain averaged probabilities
\STATE $\tilde{\boldsymbol{p}}_{\pi(1)}, ..., \tilde{\boldsymbol{p}}_{\pi(\lfloor (1-\beta)N_t \rfloor)}$ for class prediction 
\caption{Selective prediction procedure} 
\label{alg: SelectivePrediction}
\end{algorithmic}
\end{algorithm}

\begin{algorithm}[!hbtp]
\begin{algorithmic}
\STATE\textbf{Input:} ID and OOD test samples: $\{\boldsymbol{x}_{\text{id}, 1}, ..., \boldsymbol{x}_{\text{id}, N_{\text{id}}}\}$ and $\{\boldsymbol{x}_{\text{ood}, 1}, ..., \boldsymbol{x}_{\text{ood}, N_{\text{ood}}}\}$; second-order predictor $f(\cdot)$
\STATE \textbf{1.} Set labels to build the full detection test data
\STATE $\mathcal{D}_{\text{detect}}={\{(\boldsymbol{x}_{\text{id}, n}, y_n=0)\}}_{n=1}^{N_{\text{id}}}\cup {\{(\boldsymbol{x}_{\text{ood}, n}, y_n=1)\}}_{n=1}^{N_{\text{ood}}}$
\STATE \textbf{2.} Obtain second-order representations from $f(\cdot)$ on $\forall \boldsymbol{x}_n \in \mathcal{D}_{\text{detect}}$ 
\STATE $\mathcal{B}_n\leftarrow f(\boldsymbol{x}_n)$ (practical Bayesian) or $\mathcal{K}_n\leftarrow\mathcal{B}_n$ using~\eqref{Eq: constructK} (credal)
\STATE \textbf{3.} Quantify epistemic predictive uncertainty given each sample, denoted by $u_{\text{EU},n}$
\STATE $u_{\text{EU},n} \ \forall n $ using~\eqref{Eq: EUBNN_Class}/\eqref{Eq: EUBNN_Variance}/\eqref{Eq: EUBNN_WD} for $\mathcal{B}_n$ or~\eqref{Eq: EUCredal_Entropy}/\eqref{Eq: EpiUncertaintyCredal}/\eqref{Eq: MMI} for $\mathcal{K}_n$
\STATE \textbf{4.} Compute the AUROC using EU estimates for all test samples 
\STATE $\operatorname*{\text{auroc\_score}}\big((u_{\text{EU},1},..., u_{\text{EU},N_{\text{detect}}}), (y_{1},..., y_{N_{\text{detect}}})\big)$
\caption{Out-of-distribution detection procedure} 
\label{alg: ood-algorithm}
\end{algorithmic}
\end{algorithm}

\newpage
\subsection{Ship classification datasets}
\label{App: Ship}
\textbf{SeaShip dataset.}
To derive a classification dataset from existing object detection benchmarks. The target includes six classes of ships, including bulk cargo carriers, container ships, fishing boats, general cargo ships, ore carriers, and passenger ships. We are designing a cropping pipeline that transforms each annotated bounding box into an independent image crop. The procedure is summarized below:

\begin{enumerate}
\item \emph{Input format.} 
We are assuming standard YOLO-style annotations, where each object is represented as 
$(c, x_{ctr}, y_{ctr}, w, h)$ in normalized coordinates. 

\item \emph{Bounding-box expansion.} 
For each object, we are randomly expanding its bounding box by a multiplicative factor. 
The horizontal and vertical expansion ratios are being independently sampled from 
$[1+\alpha_{\min},\, 1+\alpha_{\max}]$, with $\alpha_{\min}=0.25$ and $\alpha_{\max}=0.50$ in our experiments. 
This step is ensuring that the classification model is seeing the object together with limited contextual background, 
thereby avoiding overly tight crops. 

\item \emph{Cropping and clamping.} 
The expanded box is being converted into pixel coordinates and is being clamped to the image boundaries. 
The resulting crops are being extracted and saved as individual images, each inheriting the original class label. 

\item \emph{Filtering based on area.}
The derived images are being filtered using a threshold of $256 \times 256$ pixels. 
Tiny images are being removed.

\item \emph{Manifest generation.} 
Alongside the image crops, we are generating a JSON manifest containing, for each sample, 
the file path, source image identifier, class ID, original and expanded bounding boxes 
(in COCO $(x,y,w,h)$ format), the new crop area, and the ambiguity flag. 
This manifest is enabling reproducibility and facilitating downstream training pipelines. 
\end{enumerate}

Overall, this procedure is converting every annotated object in the detection dataset into one or more classification samples, while preserving traceability to the source image and the original detection labels. The train/validation/test split is being kept consistent with the experimental settings presented in~\citet{wang2024navigating, wang2025enhancing}.

\textbf{SeaShip-C dataset.}
In addition to the clean train/validation/test classification splits described above, we are further constructing corrupted classification test sets aligned with the SeaShip-C benchmark in~\citet{wang2025enhancing}. 
SeaShip-C is defining 25 synthesized corruption types that are being applied to clean images and evaluated across multiple datasets and models. The box labels remain unchanged after corruption. 
\begin{figure}[!htbp]
\centering
\includegraphics[width=0.8\linewidth]{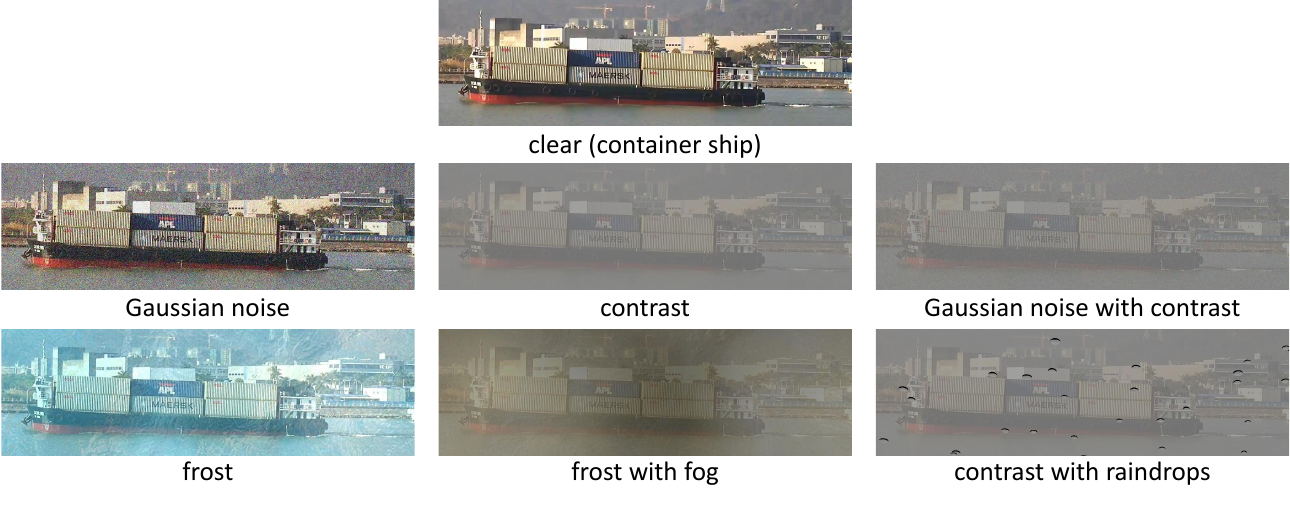}
\caption{Examples of the SeaShip dataset v.s. instances from the Seaship-C at severity level 3.}
\label{FIG: Dataset}
\end{figure}

In each experiment (i.e., a given model on a given dataset), the corruptions are being categorized as mild, moderate, or severe depending on the observed degradation in model performance. For our study, we are focusing on the subset of corruptions that are being identified as severe in at least one experimental setting reported in~\citet{wang2025enhancing}. This selection is yielding 6 corruption types: Gaussian noise, frost, contrast, Gaussian noise with contrast, contrast with raindrops, and frost with fog. For each corruption type and severity level ($\{1,3,5\}$), the classification crops are being re-extracted from the corresponding corrupted detection images, 
ensuring one-to-one alignment with the clean test set, as shown in \figurename~\ref{FIG: Dataset}.

\textbf{SeaShip-O dataset cropped from SSAVE and SMD datasets.}
In addition to the classification datasets derived from SeaShip, we are also constructing two out-of-distribution (OOD) classification datasets using the SMD and SSAVE datasets in~\citet{wang2025enhancing}. 
The procedure is following the same cropping strategy described above, using identical expansion parameters for bounding-box augmentation. 
After cropping the training partitions of both datasets, we are applying two additional steps:

\begin{enumerate}
	\item \emph{Filtering.} 
	Crops with an effective area smaller than $128 \times 128$ pixels are being discarded in order to exclude tiny instances. 
	\item \emph{Downsampling.} 
	To address the repeated appearance of vessels across frames, we are downsampling the cropped datasets: for SMD we are retaining one crop out of every 8 ($8{:}1$), and for SSAVE one out of every 4 ($4{:}1$). 
\end{enumerate}

For the label space, we are following~\citet{wang2025enhancing} and using the original class taxonomy in SMD. 
For SSAVE, we are restricting the dataset to the ship class only, since other categories are not relevant to our study. 
The resulting datasets are serving as OOD testbeds for classification, complementing the in-distribution dataset derived from SeaShip.

\newpage
\section{Additional Experimental Results}
\label{App: AddtionalResults}
\begin{table*}[!htbp]
	\centering
	\caption{AUARC scores for selective prediction tasks on various datasets.}
	\label{Table: AUARCscores}
	\small
	\begin{tabular}{@{}l|cccccc|c@{}}
		\toprule
		& DE & PackEns & MaskEns & BatchEns & SVI & MCDO & Overall \\
		\midrule \midrule 
		\multicolumn{8}{c}{In-distribution Camelyon17 test data} \\ \midrule
		MI & 0.9737{\scriptsize $\pm$0.0024} & 0.9725{\scriptsize $\pm$0.0021} & 0.9722{\scriptsize $\pm$0.0033} & 0.9742{\scriptsize $\pm$0.0033} & 0.9643{\scriptsize $\pm$0.0161} & 0.9041{\scriptsize $\pm$0.1289} & 0.9602{\scriptsize $\pm$0.0568} \\
		LWV & 0.9749{\scriptsize $\pm$0.0020} & 0.9750{\scriptsize $\pm$0.0020} & 0.9748{\scriptsize $\pm$0.0028} & 0.9765{\scriptsize $\pm$0.0025} & 0.9655{\scriptsize $\pm$0.0139} & 0.9082{\scriptsize $\pm$0.1195} & 0.9625{\scriptsize $\pm$0.0532} \\
		WD & 0.9756{\scriptsize $\pm$0.0018} & 0.9754{\scriptsize $\pm$0.0020} & 0.9753{\scriptsize $\pm$0.0028} & 0.9770{\scriptsize $\pm$0.0025} & 0.9659{\scriptsize $\pm$0.0139} & 0.9106{\scriptsize $\pm$0.1200} & 0.9633{\scriptsize $\pm$0.0530} \\ \midrule 
		GH & 0.9754{\scriptsize $\pm$0.0019} & 0.9754{\scriptsize $\pm$0.0020} & 0.9753{\scriptsize $\pm$0.0028} & 0.9769{\scriptsize $\pm$0.0026} & 0.9656{\scriptsize $\pm$0.0144} & 0.9111{\scriptsize $\pm$0.1183} & 0.9633{\scriptsize $\pm$0.0523} \\
		$H_{\mathrm{diff}}$ & 0.9726{\scriptsize $\pm$0.0020} & 0.9705{\scriptsize $\pm$0.0026} & 0.9708{\scriptsize $\pm$0.0036} & 0.9731{\scriptsize $\pm$0.0035} & 0.9616{\scriptsize $\pm$0.0179} & 0.9080{\scriptsize $\pm$0.1164} & 0.9594{\scriptsize $\pm$0.0517} \\
		MMI & 0.9754{\scriptsize $\pm$0.0019} & 0.9754{\scriptsize $\pm$0.0020} & 0.9753{\scriptsize $\pm$0.0028} & 0.9769{\scriptsize $\pm$0.0026} & 0.9656{\scriptsize $\pm$0.0144} & 0.9111{\scriptsize $\pm$0.1183} & 0.9633{\scriptsize $\pm$0.0523} \\
		\midrule \midrule 
		\multicolumn{8}{c}{Distribution-shift Camelyon17 test data} \\ \midrule
		MI & 0.9575{\scriptsize $\pm$0.0043} & 0.9686{\scriptsize $\pm$0.0039} & 0.9704{\scriptsize $\pm$0.0024} & 0.9694{\scriptsize $\pm$0.0033} & 0.9658{\scriptsize $\pm$0.0093} & 0.9507{\scriptsize $\pm$0.0394} & 0.9637{\scriptsize $\pm$0.0176} \\
		LWV & 0.9595{\scriptsize $\pm$0.0037} & 0.9715{\scriptsize $\pm$0.0040} & 0.9728{\scriptsize $\pm$0.0025} & 0.9730{\scriptsize $\pm$0.0027} & 0.9672{\scriptsize $\pm$0.0094} & 0.9573{\scriptsize $\pm$0.0271} & 0.9669{\scriptsize $\pm$0.0131} \\
		WD & 0.9620{\scriptsize $\pm$0.0034} & 0.9724{\scriptsize $\pm$0.0037} & 0.9738{\scriptsize $\pm$0.0022} & 0.9738{\scriptsize $\pm$0.0025} & 0.9685{\scriptsize $\pm$0.0089} & 0.9594{\scriptsize $\pm$0.0263} & 0.9683{\scriptsize $\pm$0.0125} \\ \midrule
		GH & 0.9614{\scriptsize $\pm$0.0036} & 0.9722{\scriptsize $\pm$0.0037} & 0.9736{\scriptsize $\pm$0.0023} & 0.9735{\scriptsize $\pm$0.0025} & 0.9683{\scriptsize $\pm$0.0089} & 0.9592{\scriptsize $\pm$0.0264} & 0.9680{\scriptsize $\pm$0.0126} \\
		$H_{\mathrm{diff}}$ & 0.9558{\scriptsize $\pm$0.0041} & 0.9663{\scriptsize $\pm$0.0032} & 0.9685{\scriptsize $\pm$0.0025} & 0.9668{\scriptsize $\pm$0.0037} & 0.9616{\scriptsize $\pm$0.0102} & 0.9490{\scriptsize $\pm$0.0359} & 0.9613{\scriptsize $\pm$0.0164} \\
		MMI & 0.9614{\scriptsize $\pm$0.0036} & 0.9722{\scriptsize $\pm$0.0037} & 0.9736{\scriptsize $\pm$0.0023} & 0.9735{\scriptsize $\pm$0.0025} & 0.9683{\scriptsize $\pm$0.0089} & 0.9592{\scriptsize $\pm$0.0264} & 0.9680{\scriptsize $\pm$0.0126} \\
		\midrule \midrule 
		\multicolumn{8}{c}{In-distribution SeaShip test data} \\ \midrule
		MI & 0.9962{\scriptsize $\pm$0.0006} & 0.9952{\scriptsize $\pm$0.0015} & 0.9931{\scriptsize $\pm$0.0018} & 0.9929{\scriptsize $\pm$0.0023} & 0.9639{\scriptsize $\pm$0.0267} & 0.9910{\scriptsize $\pm$0.0058} & 0.9887{\scriptsize $\pm$0.0156} \\
		LWV & 0.9968{\scriptsize $\pm$0.0005} & 0.9962{\scriptsize $\pm$0.0011} & 0.9941{\scriptsize $\pm$0.0019} & 0.9939{\scriptsize $\pm$0.0022} & 0.9660{\scriptsize $\pm$0.0260} & 0.9922{\scriptsize $\pm$0.0053} & 0.9899{\scriptsize $\pm$0.0151} \\
		WD & 0.9971{\scriptsize $\pm$0.0005} & 0.9964{\scriptsize $\pm$0.0010} & 0.9945{\scriptsize $\pm$0.0017} & 0.9943{\scriptsize $\pm$0.0020} & 0.9676{\scriptsize $\pm$0.0249} & 0.9924{\scriptsize $\pm$0.0051} & 0.9904{\scriptsize $\pm$0.0144} \\ \midrule
		GH & 0.9970{\scriptsize $\pm$0.0005} & 0.9963{\scriptsize $\pm$0.0011} & 0.9944{\scriptsize $\pm$0.0016} & 0.9943{\scriptsize $\pm$0.0020} & 0.9675{\scriptsize $\pm$0.0248} & 0.9924{\scriptsize $\pm$0.0051} & 0.9903{\scriptsize $\pm$0.0144} \\
		$H_{\mathrm{diff}}$ & 0.9947{\scriptsize $\pm$0.0005} & 0.9942{\scriptsize $\pm$0.0019} & 0.9917{\scriptsize $\pm$0.0022} & 0.9914{\scriptsize $\pm$0.0031} & 0.9479{\scriptsize $\pm$0.0423} & 0.9885{\scriptsize $\pm$0.0070} & 0.9848{\scriptsize $\pm$0.0237} \\
		MMI & 0.9969{\scriptsize $\pm$0.0004} & 0.9963{\scriptsize $\pm$0.0011} & 0.9943{\scriptsize $\pm$0.0017} & 0.9941{\scriptsize $\pm$0.0021} & 0.9667{\scriptsize $\pm$0.0259} & 0.9924{\scriptsize $\pm$0.0051} & 0.9901{\scriptsize $\pm$0.0149} \\
		\midrule \midrule 
		\multicolumn{8}{c}{In-distribution CIFAR10 test data} \\ \midrule
		MI & 0.9828{\scriptsize $\pm$0.0013} & 0.9784{\scriptsize $\pm$0.0029} & 0.9851{\scriptsize $\pm$0.0018} & 0.9823{\scriptsize $\pm$0.0039} & 0.9763{\scriptsize $\pm$0.0078} & 0.9794{\scriptsize $\pm$0.0038} & 0.9807{\scriptsize $\pm$0.0050} \\
		LWV & 0.9836{\scriptsize $\pm$0.0011} & 0.9796{\scriptsize $\pm$0.0027} & 0.9855{\scriptsize $\pm$0.0016} & 0.9832{\scriptsize $\pm$0.0035} & 0.9767{\scriptsize $\pm$0.0078} & 0.9801{\scriptsize $\pm$0.0035} & 0.9815{\scriptsize $\pm$0.0049} \\
		WD & 0.9845{\scriptsize $\pm$0.0012} & 0.9805{\scriptsize $\pm$0.0026} & 0.9864{\scriptsize $\pm$0.0015} & 0.9840{\scriptsize $\pm$0.0033} & 0.9777{\scriptsize $\pm$0.0072} & 0.9810{\scriptsize $\pm$0.0034} & 0.9823{\scriptsize $\pm$0.0046} \\ \midrule
		GH & 0.9843{\scriptsize $\pm$0.0011} & 0.9805{\scriptsize $\pm$0.0025} & 0.9863{\scriptsize $\pm$0.0016} & 0.9839{\scriptsize $\pm$0.0034} & 0.9776{\scriptsize $\pm$0.0072} & 0.9807{\scriptsize $\pm$0.0034} & 0.9822{\scriptsize $\pm$0.0046} \\
		$H_{\mathrm{diff}}$ & 0.9790{\scriptsize $\pm$0.0020} & 0.9734{\scriptsize $\pm$0.0041} & 0.9823{\scriptsize $\pm$0.0026} & 0.9791{\scriptsize $\pm$0.0052} & 0.9707{\scriptsize $\pm$0.0105} & 0.9739{\scriptsize $\pm$0.0057} & 0.9764{\scriptsize $\pm$0.0068} \\
		MMI & 0.9840{\scriptsize $\pm$0.0011} & 0.9801{\scriptsize $\pm$0.0026} & 0.9860{\scriptsize $\pm$0.0016} & 0.9837{\scriptsize $\pm$0.0034} & 0.9771{\scriptsize $\pm$0.0074} & 0.9803{\scriptsize $\pm$0.0035} & 0.9819{\scriptsize $\pm$0.0048} \\
		\bottomrule
	\end{tabular}
\end{table*}
\begin{table*}[!htbp]
\centering
\caption{AUROC scores for OOD detection tasks on various dataset pairs.}
\label{Table: AUROCscores}
\small
\begin{tabular}{@{}l|cccccc|c@{}}
\toprule
& DE & PackEns & MaskEns & BatchEns & SVI & MCDO & Overall \\
\midrule \midrule 
\multicolumn{8}{c}{SeaShip v.s. SeaShip-O} \\ \midrule
MI & 0.9153{\scriptsize $\pm$0.0026} & 0.9347{\scriptsize $\pm$0.0127} & 0.9145{\scriptsize $\pm$0.0106} & 0.9072{\scriptsize $\pm$0.0139} & 0.7129{\scriptsize $\pm$0.0678} & 0.8315{\scriptsize $\pm$0.0379} & 0.8694{\scriptsize $\pm$0.0840} \\
LWV & 0.9085{\scriptsize $\pm$0.0021} & 0.9265{\scriptsize $\pm$0.0136} & 0.9066{\scriptsize $\pm$0.0107} & 0.8986{\scriptsize $\pm$0.0139} & 0.7147{\scriptsize $\pm$0.0672} & 0.8378{\scriptsize $\pm$0.0355} & 0.8655{\scriptsize $\pm$0.0797} \\
WD & 0.9235{\scriptsize $\pm$0.0028} & 0.9373{\scriptsize $\pm$0.0136} & 0.9202{\scriptsize $\pm$0.0108} & 0.9129{\scriptsize $\pm$0.0128} & 0.7449{\scriptsize $\pm$0.0681} & 0.8545{\scriptsize $\pm$0.0322} & 0.8822{\scriptsize $\pm$0.0739} \\ \midrule
GH & 0.9240{\scriptsize $\pm$0.0025} & 0.9403{\scriptsize $\pm$0.0125} & 0.9235{\scriptsize $\pm$0.0104} & 0.9161{\scriptsize $\pm$0.0126} & 0.7519{\scriptsize $\pm$0.0654} & 0.8577{\scriptsize $\pm$0.0314} & 0.8856{\scriptsize $\pm$0.0720} \\
$H_{\mathrm{diff}}$ & 0.9023{\scriptsize $\pm$0.0077} & 0.9220{\scriptsize $\pm$0.0159} & 0.8939{\scriptsize $\pm$0.0136} & 0.8913{\scriptsize $\pm$0.0174} & 0.6841{\scriptsize $\pm$0.0725} & 0.8162{\scriptsize $\pm$0.0397} & 0.8516{\scriptsize $\pm$0.0893} \\
MMI & 0.9168{\scriptsize $\pm$0.0024} & 0.9339{\scriptsize $\pm$0.0131} & 0.9159{\scriptsize $\pm$0.0104} & 0.9086{\scriptsize $\pm$0.0136} & 0.7356{\scriptsize $\pm$0.0656} & 0.8503{\scriptsize $\pm$0.0323} & 0.8768{\scriptsize $\pm$0.0751} \\
\midrule \midrule 
\multicolumn{8}{c}{SeaShip v.s. SeaShip-C} \\ \midrule
MI & 0.8815{\scriptsize $\pm$0.0063} & 0.8892{\scriptsize $\pm$0.0146} & 0.8709{\scriptsize $\pm$0.0101} & 0.8720{\scriptsize $\pm$0.0141} & 0.6734{\scriptsize $\pm$0.0576} & 0.7629{\scriptsize $\pm$0.0414} & 0.8250{\scriptsize $\pm$0.0859} \\
LWV & 0.8782{\scriptsize $\pm$0.0061} & 0.8868{\scriptsize $\pm$0.0139} & 0.8696{\scriptsize $\pm$0.0101} & 0.8696{\scriptsize $\pm$0.0137} & 0.6870{\scriptsize $\pm$0.0565} & 0.7886{\scriptsize $\pm$0.0382} & 0.8300{\scriptsize $\pm$0.0776} \\
WD & 0.8990{\scriptsize $\pm$0.0048} & 0.9033{\scriptsize $\pm$0.0128} & 0.8864{\scriptsize $\pm$0.0080} & 0.8898{\scriptsize $\pm$0.0112} & 0.7214{\scriptsize $\pm$0.0534} & 0.8071{\scriptsize $\pm$0.0365} & 0.8512{\scriptsize $\pm$0.0720} \\\midrule
GH & 0.9028{\scriptsize $\pm$0.0050} & 0.9077{\scriptsize $\pm$0.0114} & 0.8908{\scriptsize $\pm$0.0074} & 0.8943{\scriptsize $\pm$0.0103} & 0.7299{\scriptsize $\pm$0.0495} & 0.8110{\scriptsize $\pm$0.0361} & 0.8561{\scriptsize $\pm$0.0703} \\
$H_{\mathrm{diff}}$ & 0.8496{\scriptsize $\pm$0.0088} & 0.8632{\scriptsize $\pm$0.0195} & 0.8418{\scriptsize $\pm$0.0121} & 0.8353{\scriptsize $\pm$0.0216} & 0.6248{\scriptsize $\pm$0.0626} & 0.7425{\scriptsize $\pm$0.0400} & 0.7929{\scriptsize $\pm$0.0912} \\
MMI & 0.8926{\scriptsize $\pm$0.0058} & 0.8993{\scriptsize $\pm$0.0124} & 0.8827{\scriptsize $\pm$0.0082} & 0.8848{\scriptsize $\pm$0.0119} & 0.7128{\scriptsize $\pm$0.0514} & 0.8036{\scriptsize $\pm$0.0365} & 0.8460{\scriptsize $\pm$0.0729} \\
\midrule \midrule 
\multicolumn{8}{c}{CIFAR10 v.s. SVHN} \\ \midrule
MI & 0.7818{\scriptsize $\pm$0.0094} & 0.7561{\scriptsize $\pm$0.0119} & 0.7940{\scriptsize $\pm$0.0159} & 0.7894{\scriptsize $\pm$0.0200} & 0.7435{\scriptsize $\pm$0.0272} & 0.7292{\scriptsize $\pm$0.0192} & 0.7657{\scriptsize $\pm$0.0301} \\
LWV & 0.7806{\scriptsize $\pm$0.0099} & 0.7557{\scriptsize $\pm$0.0136} & 0.7968{\scriptsize $\pm$0.0147} & 0.7927{\scriptsize $\pm$0.0214} & 0.7467{\scriptsize $\pm$0.0346} & 0.7440{\scriptsize $\pm$0.0221} & 0.7694{\scriptsize $\pm$0.0295} \\
WD & 0.8326{\scriptsize $\pm$0.0082} & 0.8046{\scriptsize $\pm$0.0101} & 0.8366{\scriptsize $\pm$0.0128} & 0.8321{\scriptsize $\pm$0.0168} & 0.8043{\scriptsize $\pm$0.0246} & 0.7944{\scriptsize $\pm$0.0168} & 0.8174{\scriptsize $\pm$0.0227} \\ \midrule
GH & 0.8445{\scriptsize $\pm$0.0073} & 0.8191{\scriptsize $\pm$0.0081} & 0.8463{\scriptsize $\pm$0.0120} & 0.8435{\scriptsize $\pm$0.0145} & 0.8218{\scriptsize $\pm$0.0189} & 0.8068{\scriptsize $\pm$0.0134} & 0.8303{\scriptsize $\pm$0.0197} \\
$H_{\mathrm{diff}}$ & 0.7950{\scriptsize $\pm$0.0132} & 0.7579{\scriptsize $\pm$0.0192} & 0.8121{\scriptsize $\pm$0.0214} & 0.8002{\scriptsize $\pm$0.0270} & 0.7622{\scriptsize $\pm$0.0397} & 0.7509{\scriptsize $\pm$0.0268} & 0.7797{\scriptsize $\pm$0.0343} \\
MMI & 0.8150{\scriptsize $\pm$0.0081} & 0.7927{\scriptsize $\pm$0.0103} & 0.8246{\scriptsize $\pm$0.0124} & 0.8229{\scriptsize $\pm$0.0167} & 0.7883{\scriptsize $\pm$0.0255} & 0.7795{\scriptsize $\pm$0.0169} & 0.8038{\scriptsize $\pm$0.0235} \\
\midrule \midrule 
\multicolumn{8}{c}{CIFAR10 v.s. FMNIST} \\ \midrule
MI & 0.8852{\scriptsize $\pm$0.0097} & 0.8675{\scriptsize $\pm$0.0096} & 0.8929{\scriptsize $\pm$0.0111} & 0.8740{\scriptsize $\pm$0.0105} & 0.8271{\scriptsize $\pm$0.0231} & 0.8366{\scriptsize $\pm$0.0117} & 0.8639{\scriptsize $\pm$0.0276} \\
LWV & 0.8653{\scriptsize $\pm$0.0103} & 0.8456{\scriptsize $\pm$0.0101} & 0.8736{\scriptsize $\pm$0.0109} & 0.8580{\scriptsize $\pm$0.0123} & 0.8149{\scriptsize $\pm$0.0231} & 0.8224{\scriptsize $\pm$0.0134} & 0.8466{\scriptsize $\pm$0.0256} \\
WD & 0.8907{\scriptsize $\pm$0.0082} & 0.8736{\scriptsize $\pm$0.0084} & 0.8992{\scriptsize $\pm$0.0098} & 0.8804{\scriptsize $\pm$0.0103} & 0.8438{\scriptsize $\pm$0.0204} & 0.8508{\scriptsize $\pm$0.0105} & 0.8731{\scriptsize $\pm$0.0232} \\ \midrule
GH & 0.8968{\scriptsize $\pm$0.0075} & 0.8816{\scriptsize $\pm$0.0079} & 0.9059{\scriptsize $\pm$0.0098} & 0.8867{\scriptsize $\pm$0.0089} & 0.8512{\scriptsize $\pm$0.0200} & 0.8584{\scriptsize $\pm$0.0096} & 0.8801{\scriptsize $\pm$0.0226} \\
$H_{\mathrm{diff}}$ & 0.8912{\scriptsize $\pm$0.0096} & 0.8716{\scriptsize $\pm$0.0128} & 0.9043{\scriptsize $\pm$0.0120} & 0.8835{\scriptsize $\pm$0.0118} & 0.8416{\scriptsize $\pm$0.0286} & 0.8542{\scriptsize $\pm$0.0154} & 0.8744{\scriptsize $\pm$0.0266} \\
MMI & 0.8799{\scriptsize $\pm$0.0087} & 0.8644{\scriptsize $\pm$0.0085} & 0.8885{\scriptsize $\pm$0.0100} & 0.8732{\scriptsize $\pm$0.0096} & 0.8322{\scriptsize $\pm$0.0208} & 0.8400{\scriptsize $\pm$0.0106} & 0.8630{\scriptsize $\pm$0.0237} \\
\bottomrule
\end{tabular}
\end{table*}
\begin{table}[!htbp]
\centering
\small
\setlength\tabcolsep{4pt}
\caption{Net wins for intra-representation comparisons on selective prediction tasks across datasets. The first rank according to net wins is shown in bold red, and the second rank in bold blue.}
\label{Table: NetWinsIntraSP}
\begin{tabular}{@{}lccccccc|ccccccc@{}}
\toprule
 & DE & PackEns & MaskEns & BatchEns & SVI & MCDO & Total & DE & PackEns & MaskEns & BatchEns & SVI & MCDO & Total \\
\midrule \midrule
& \multicolumn{14}{c}{Distribution-based measures} \\ \midrule
 & \multicolumn{7}{c|}{In-distribution Camelyon17 test data} &\multicolumn{7}{c}{Distribution-shift Camelyon17 test data} \\ \midrule 
MI & -2 & -2 & -2 & -2 & -1 & -2 & -11 & -2 & -2 & -2 & -2 & -2 & -2 & -12 \\
LWV & 0 & 0 & 0 & 0 & 0 & 0 & \color{blue}\textbf{0} & 0 & 0 & 0 & 0 & 0 & 0 & \color{blue}\textbf{0}\\
WD & 2 & 2 & 2 & 2 & 1 & 2 & \color{red}\textbf{11}  & 2 & 2 & 2 & 2 & 2 & 2 & \color{red}\textbf{12}\\ \midrule
 & \multicolumn{7}{c|}{In-distribution SeaShip test data} &\multicolumn{7}{c}{In-distribution CIFAR10 test data} \\ \midrule
MI & -2 & -2 & -2 & -2 & -2 & -2 & -12 & -2 & -2 & -2 & -2 & -2 & -2 & -12 \\
LWV & 0 & 0 & 0 & 0 & 0 & 0 & \color{blue}\textbf{0} & 0 & 0 & 0 & 0 & 0 & 0 & \color{blue}\textbf{0}\\
WD & 2 & 2 & 2 & 2 & 2 & 2 & \color{red}\textbf{12} & 2 & 2 & 2 & 2 & 2 & 2 & \color{red}\textbf{12}\\ \midrule \midrule 
& \multicolumn{14}{c}{Credal-based measures} \\ \midrule
 & \multicolumn{7}{c|}{In-distribution Camelyon17 test data} &\multicolumn{7}{c}{Distribution-shift Camelyon17 test data} \\ \midrule
GH & 1 & 1 & 1 & 1 & 1 & 1 & \textbf{\textcolor{red}{6}} & 1 & 1 & 1 & 1 & 1 & 1 & \textbf{\textcolor{red}{6}} \\
$H_{\mathrm{diff}}$ & -2 & -2 & -2 & -2 & -2 & -2 & -12 & -2 & -2 & -2 & -2 & -2 & -2 & -12\\
MMI & 1 & 1 & 1 & 1 & 1 & 1 & \textbf{\textcolor{red}{6}}  & 1 & 1 & 1 & 1 & 1 & 1 & \textbf{\textcolor{red}{6}}\\ \midrule
 & \multicolumn{7}{c|}{In-distribution SeaShip test data} &\multicolumn{7}{c}{In-distribution CIFAR10 test data} \\ \midrule
GH & 2 & 1 & 2 & 2 & 2 & 1 & \color{red}\textbf{10} & 2 & 2 & 2 & 2 & 2 & 2 & \textbf{\textcolor{red}{12}}  \\
$H_{\mathrm{diff}}$ & -2 & -2 & -2 & -2 & -2 & -2 & -12 & -2 & -2 & -2 & -2 & -2 & -2 & -12\\
MMI & 0 & 1 & 0 & 0 & 0 & 1 & \color{blue}\textbf{2}  & 0 & 0 & 0 & 0 & 0 & 0 & \textbf{\textcolor{blue}{0}}\\
\bottomrule
\end{tabular}
\end{table}
\begin{table}[!htbp]
\centering
\small
\setlength\tabcolsep{4pt}
\caption{Net wins for inter-representation comparisons on selective prediction tasks across datasets. The first rank according to net wins is shown in bold red, and the second rank in bold blue.}
\label{Table: NetWinsInterSP}
\begin{tabular}{@{}lccccccc|ccccccc@{}}
\toprule
 & DE & PackEns & MaskEns & BatchEns & SVI & MCDO & Total & DE & PackEns & MaskEns & BatchEns & SVI & MCDO & Total \\
\midrule
 & \multicolumn{7}{c|}{In-distribution Camelyon17 test data} &\multicolumn{7}{c}{Distribution-shift Camelyon17 test data} \\ \midrule
MI & -3 & -3 & -3 & -3 & -2 & -4 & -18 & -3 & -3 & -3 & -3 & -3 & -4 & -19 \\
LWV & -1 & -1 & -1 & -1 & 1 & -2 & -5 & -1 & -1 & -1 & -1 & -1 & -1 & -6 \\
WD & 5 & 3 & 3 & 5 & 2 & 3 & \color{red}\textbf{21} & 5 & 5 & 5 & 5 & 5 & 5 & \color{red}\textbf{30} \\
GH & 2 & 3 & 3 & 2 & 2 & 3 & \color{blue}\textbf{15} & 2 & 2 & 2 & 2 & 2 & 2 & \color{blue}\textbf{12} \\
$H_{\mathrm{diff}}$ & -5 & -5 & -5 & -5 & -5 & -3 & -28 & -5 & -5 & -5 & -5 & -5 & -4 & -29 \\
MMI & 2 & 3 & 3 & 2 & 2 & 3 & \color{blue}\textbf{15}& 2 & 2 & 2 & 2 & 2 & 2 & \color{blue}\textbf{12} \\
\midrule
 & \multicolumn{7}{c|}{In-distribution SeaShip test data} &\multicolumn{7}{c}{In-distribution CIFAR10 test data} \\ \midrule
MI & -3 & -3 & -3 & -3 & -3 & -3 & -18 & -3 & -3 & -3 & -3 & -3 & -3 & -18 \\
LWV & 0 & -1 & -1 & -1 & -1 & 0 & -4 & -1 & -1 & -1 & -1 & -1 & -1 & -6 \\
WD & 5 & 5 & 4 & 4 & 4 & 3 & \color{red}\textbf{25} & 5 & 4 & 5 & 4 & 5 & 5 & \color{red}\textbf{28} \\
GH & 3 & 2 & 4 & 4 & 4 & 2 & \color{blue}\textbf{19} & 3 & 4 & 3 & 4 & 3 & 3 & \color{blue}\textbf{20} \\
$H_{\mathrm{diff}}$ & -5 & -5 & -5 & -5 & -5 & -5 & -30 & -5 & -5 & -5 & -5 & -5 & -5 & -30 \\
MMI & 0 & 2 & 1 & 1 & 1 & 3 & 8 & 1 & 1 & 1 & 1 & 1 & 1 & 6 \\
\bottomrule
\end{tabular}
\end{table}

\begin{table}[!htbp]
\centering
\small
\setlength\tabcolsep{4pt}
\caption{Net wins for intra-representation comparisons on OOD detection tasks across datasets. The first rank according to net wins is shown in bold red, and the second rank in bold blue.}
\label{Table: NetWinsIntraOOD}
\begin{tabular}{@{}lccccccc|ccccccc@{}}
\toprule
 & DE & PackEns & MaskEns & BatchEns & SVI & MCDO & Total & DE & PackEns & MaskEns & BatchEns & SVI & MCDO & Total \\
\midrule \midrule
& \multicolumn{14}{c}{Distribution-based measures} \\ \midrule
 & \multicolumn{7}{c|}{SeaShip v.s. SeaShip-O} &\multicolumn{7}{c}{SeaShip v.s. SeaShip-C} \\ \midrule 
MI & 0 & 0 & 0 & 0 & -1 & -2 & \color{blue}\textbf{-3} & 0 & 0 & -1 & 0 & -2 & -2 & \color{blue}\textbf{-5} \\
LWV & -2 & -2 & -2 & -2 & -1 & 0 & -9 & -2 & -2 & -1 & -2 & 0 & 0 & -7 \\
WD & 2 & 2 & 2 & 2 & 2 & 2 & \color{red}\textbf{12} & 2 & 2 & 2 & 2 & 2 & 2 & \color{red}\textbf{12} \\
\midrule
 & \multicolumn{7}{c|}{CIFAR10 v.s. SVHN} &\multicolumn{7}{c}{CIFAR10 v.s. FMNIST} \\ \midrule
MI & 0 & -1 & -2 & -2 & -1 & -2 & -8 & 0 & 0 & 0 & 0 & 0 & 0 & \color{blue}\textbf{0} \\
LWV & -2 & -1 & 0 & 0 & -1 & 0 & \color{blue}\textbf{-4} & -2 & -2 & -2 & -2 & -2 & -2 & -12 \\
WD & 2 & 2 & 2 & 2 & 2 & 2 & \color{red}\textbf{12} & 2 & 2 & 2 & 2 & 2 & 2 & \color{red}\textbf{12} \\
\midrule \midrule 
& \multicolumn{14}{c}{Credal-based measures} \\ \midrule
 & \multicolumn{7}{c|}{SeaShip v.s. SeaShip-O} &\multicolumn{7}{c}{SeaShip v.s. SeaShip-C} \\ \midrule
GH & 2 & 2 & 2 & 2 & 2 & 2 & \color{red}\textbf{12}  & 2 & 2 & 2 & 2 & 2 & 2 & \color{red}\textbf{12} \\
$H_{\mathrm{diff}}$ & -2 & -2 & -2 & -2 & -2 & -2 & -12 & -2 & -2 & -2 & -2 & -2 & -2 & -12 \\
MMI & 0 & 0 & 0 & 0 & 0 & 0 & \color{blue}\textbf{0} & 0 & 0 & 0 & 0 & 0 & 0 & \color{blue}\textbf{0} \\
\midrule
 & \multicolumn{7}{c|}{CIFAR10 v.s. SVHN} &\multicolumn{7}{c}{CIFAR10 v.s. FMNIST} \\ \midrule
GH & 2 & 2 & 2 & 2 & 2 & 2 & \color{red}\textbf{12} & 2 & 2 & 1 & 1 & 2 & 1 & \color{red}\textbf{9} \\
$H_{\mathrm{diff}}$ & -2 & -2 & -2 & -2 & -2 & -2 & -12 & 0 & 0 & 1 & 1 & 0 & 1 & \color{blue}\textbf{3} \\
MMI & 0 & 0 & 0 & 0 & 0 & 0 & \color{blue}\textbf{0}& -2 & -2 & -2 & -2 & -2 & -2 & -12 \\
\bottomrule
\end{tabular}
\end{table}

\begin{table}[!htbp]
\centering
\small
\setlength\tabcolsep{4pt}
\caption{Net wins for inter-representation comparisons on OOD detection tasks across datasets. The first rank according to net wins is shown in bold red, and the second rank in bold blue.}
\label{Table: NetWinsInterOOD}
\begin{tabular}{@{}lccccccc|ccccccc@{}}
\toprule
 & DE & PackEns & MaskEns & BatchEns & SVI & MCDO & Total & DE & PackEns & MaskEns & BatchEns & SVI & MCDO & Total \\
\midrule
 & \multicolumn{7}{c|}{SeaShip v.s. SeaShip-O} &\multicolumn{7}{c}{SeaShip v.s. SeaShip-C} \\ \midrule
MI & -1 & 0 & 0 & 0 & -2 & -3 & -6 & -1 & -1 & -2 & -1 & -3 & -3 & -11 \\
LWV & -3 & -3 & -3 & -3 & -2 & -1 & -15 & -3 & -3 & -2 & -3 & -1 & -1 & -13 \\
WD & 4 & 3 & 3 & 3 & 3 & 3 & \color{blue}\textbf{19} & 3 & 3 & 3 & 3 & 3 & 3 & \color{blue}\textbf{18} \\
GH & 4 & 5 & 5 & 5 & 5 & 5 & \color{red}\textbf{29} & 5 & 5 & 5 & 5 & 5 & 5 & \color{red}\textbf{30} \\
$H_{\mathrm{diff}}$ & -5 & -5 & -5 & -5 & -5 & -5 & -30 & -5 & -5 & -5 & -5 & -5 & -5 & -30 \\
MMI & 1 & 0 & 0 & 0 & 1 & 1 & 3 & 1 & 1 & 1 & 1 & 1 & 1 & 6 \\
\midrule
 & \multicolumn{7}{c|}{CIFAR10 v.s. SVHN} &\multicolumn{7}{c}{CIFAR10 v.s. FMNIST} \\ \midrule
MI & -3 & -3 & -5 & -5 & -4 & -5 & -25 & -1 & -1 & -1 & -2 & -3 & -3 & -11 \\
LWV & -5 & -3 & -3 & -3 & -4 & -3 & -21  & -5 & -5 & -5 & -5 & -5 & -5 & -30 \\
WD & 3 & 3 & 3 & 3 & 3 & 3 & \color{blue}\textbf{18} & 2 & 2 & 1 & 1 & 2 & 2 & 10 \\
GH & 5 & 5 & 5 & 5 & 5 & 5 & \color{red}\textbf{30} & 5 & 5 & 4 & 4 & 5 & 4 & \color{red}\textbf{27} \\
$H_{\mathrm{diff}}$ & -1 & -3 & -1 & -1 & -1 & -1 & -8 & 2 & 2 & 4 & 4 & 2 & 3 & \color{blue}\textbf{17} \\
MMI & 1 & 1 & 1 & 1 & 1 & 1 & 6 & -3 & -3 & -3 & -2 & -1 & -1 & -13 \\
\bottomrule
\end{tabular}
\end{table}
\begin{figure}[!htbp]
\centering
\includegraphics[width=\linewidth]{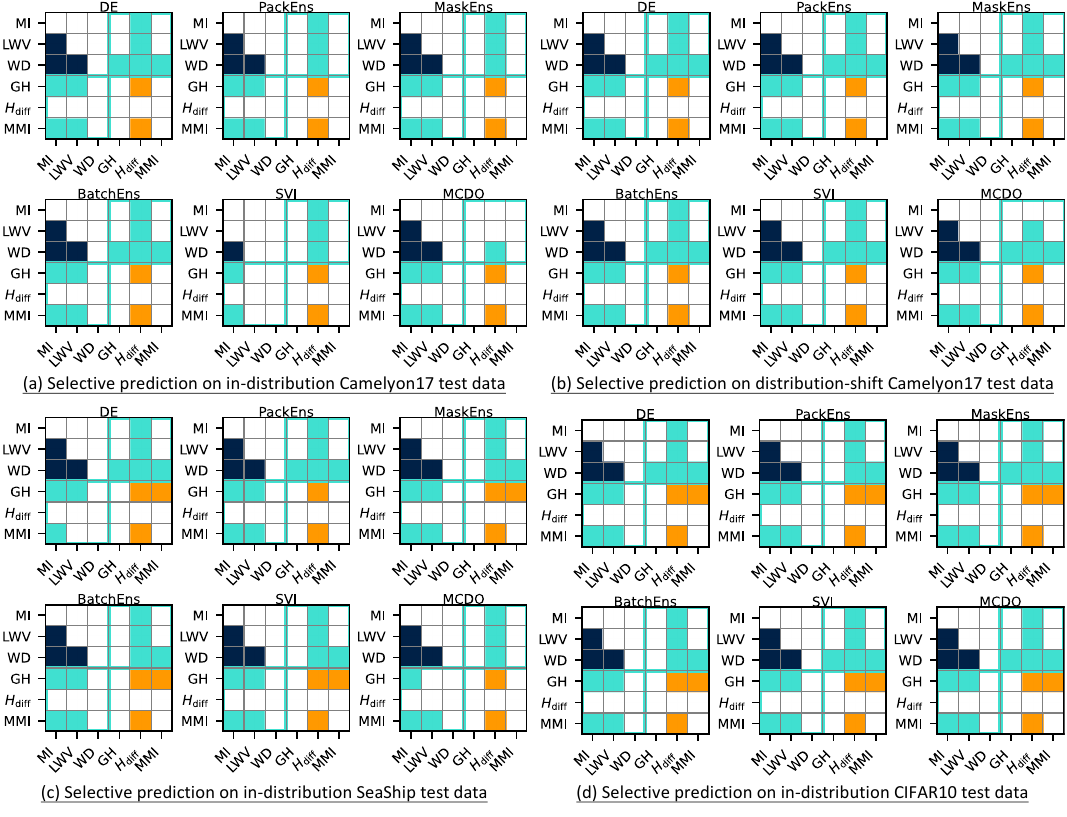}
\caption{Statistical significance plots on different selective prediction (a-d) benchmarks across different underlying predictive models. A cell is shaded if the measure in the $i$-th row is statistically significantly better than that in the $j$-th column according to a pairwise one-sided Wilcoxon signed-rank test at the 5\% significance level. Intra-representation comparisons are shown in blue (distribution-based measures) and orange (credal-based measures), while inter-representation comparisons are shown in green.}
\label{FIG: OODresults}
\end{figure}
\begin{figure}[!htbp]
\centering
\includegraphics[width=\linewidth]{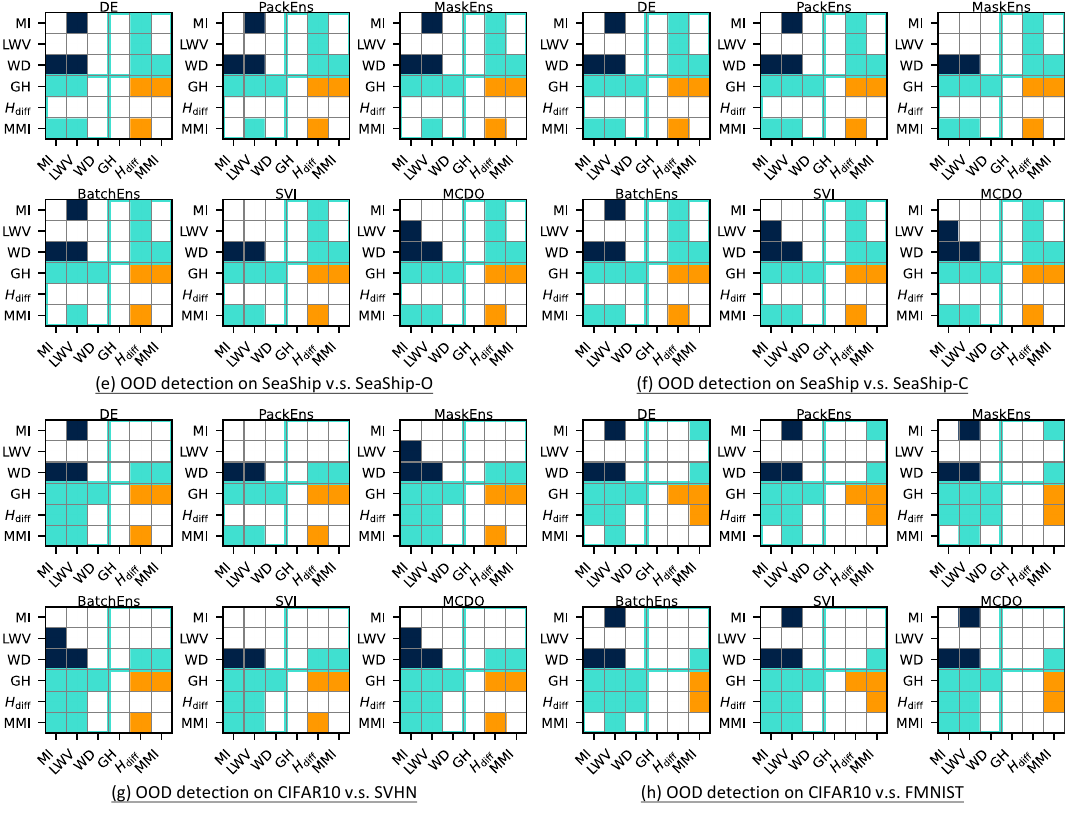}
\caption{Statistical significance plots on different OOD detection (e-h) benchmarks across different underlying predictive models. A cell is shaded if the measure in the $i$-th row is statistically significantly better than that in the $j$-th column according to a pairwise one-sided Wilcoxon signed-rank test at the 5\% significance level. Intra-representation comparisons are shown in blue (distribution-based measures) and orange (credal-based measures), while inter-representation comparisons are shown in green.}
\label{FIG: SPresults}
\end{figure}
\begin{figure}[!htbp]
\centering
\includegraphics[width=\linewidth]{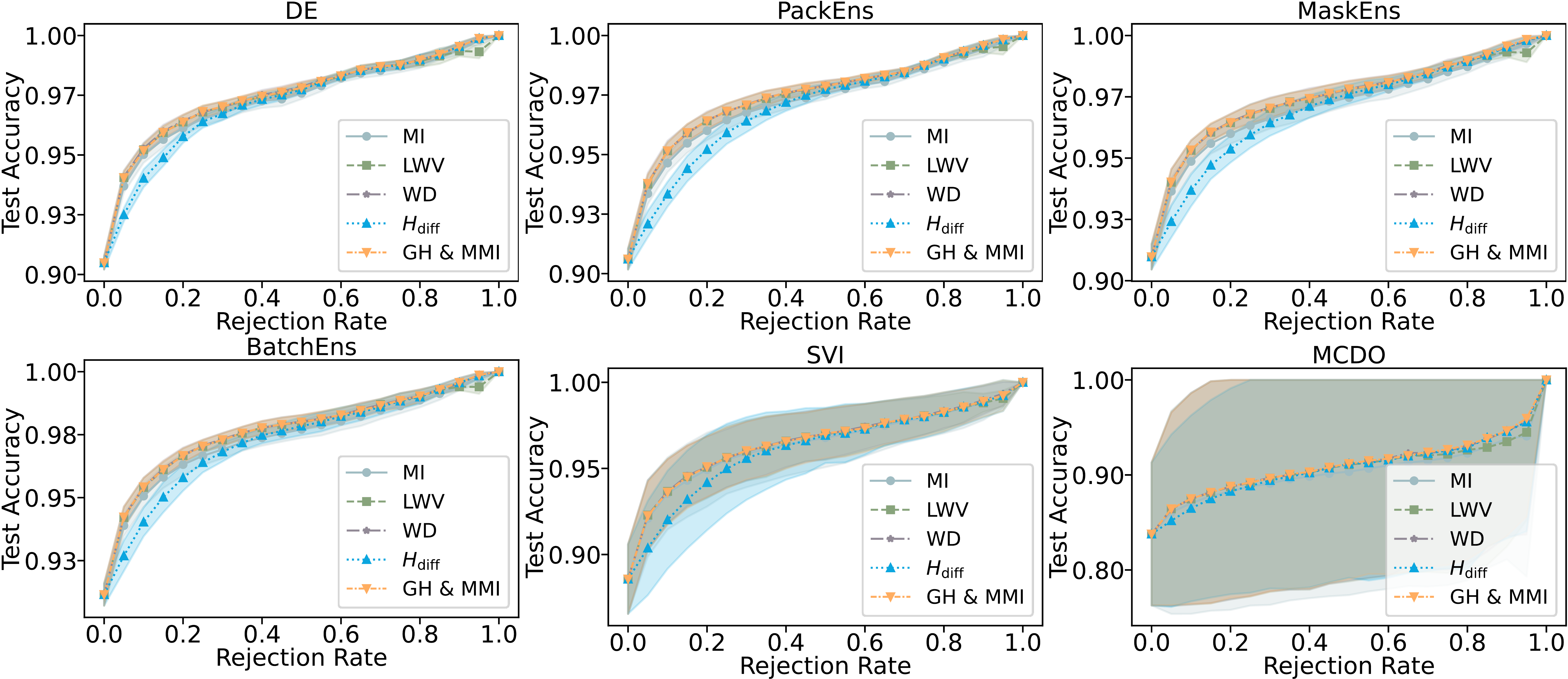}
\caption{Accuracy rejection curves using different uncertainty representations and measures on in-distribution Camelyon17 test data across different underlying predictive models.}
\label{FIG: MedicalIDARC}
\end{figure}
\begin{figure}[!htbp]
\centering
\includegraphics[width=\linewidth]{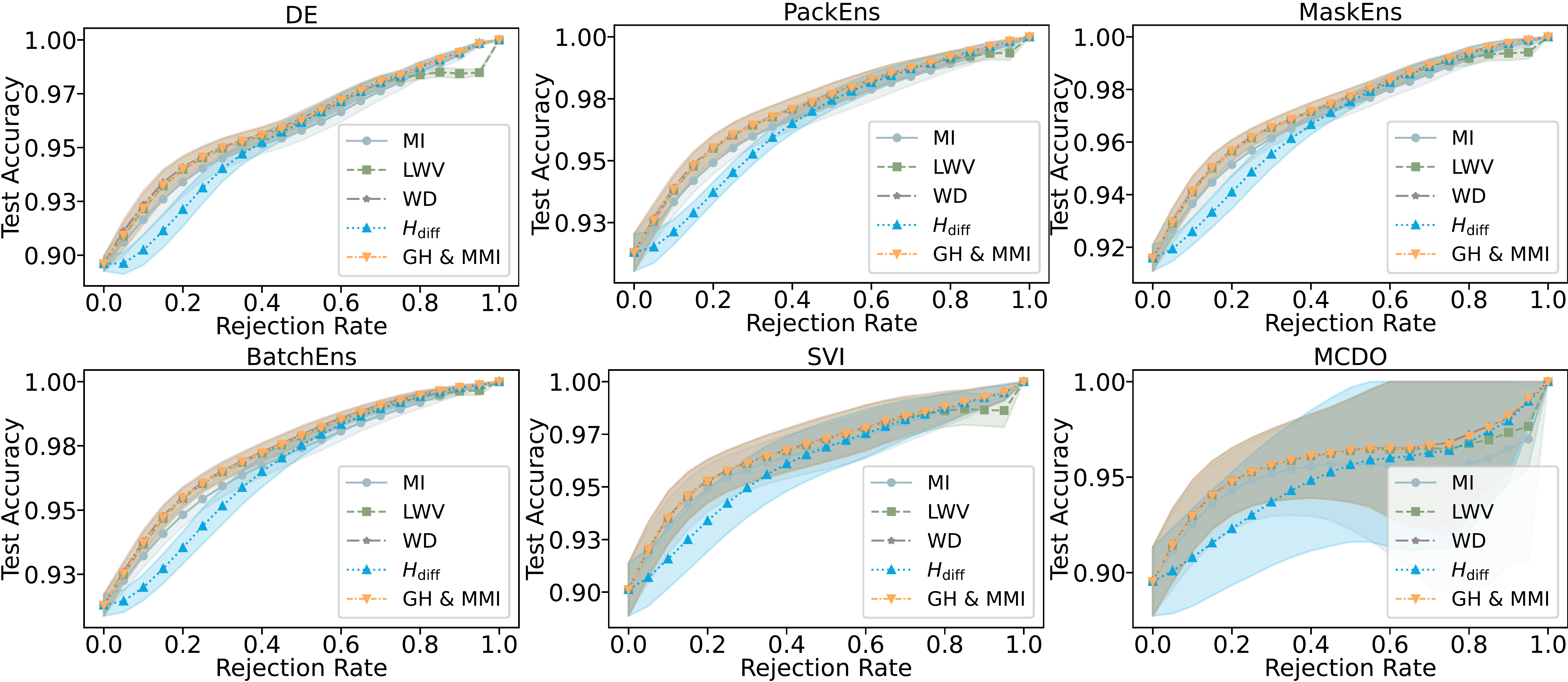}
\caption{Accuracy rejection curves using different uncertainty representations and measures on distribution-shift Camelyon17 test data across different underlying predictive models.}
\label{FIG: MedicalOODARC}
\end{figure}
\begin{figure}[!htbp]
\centering
\includegraphics[width=\linewidth]{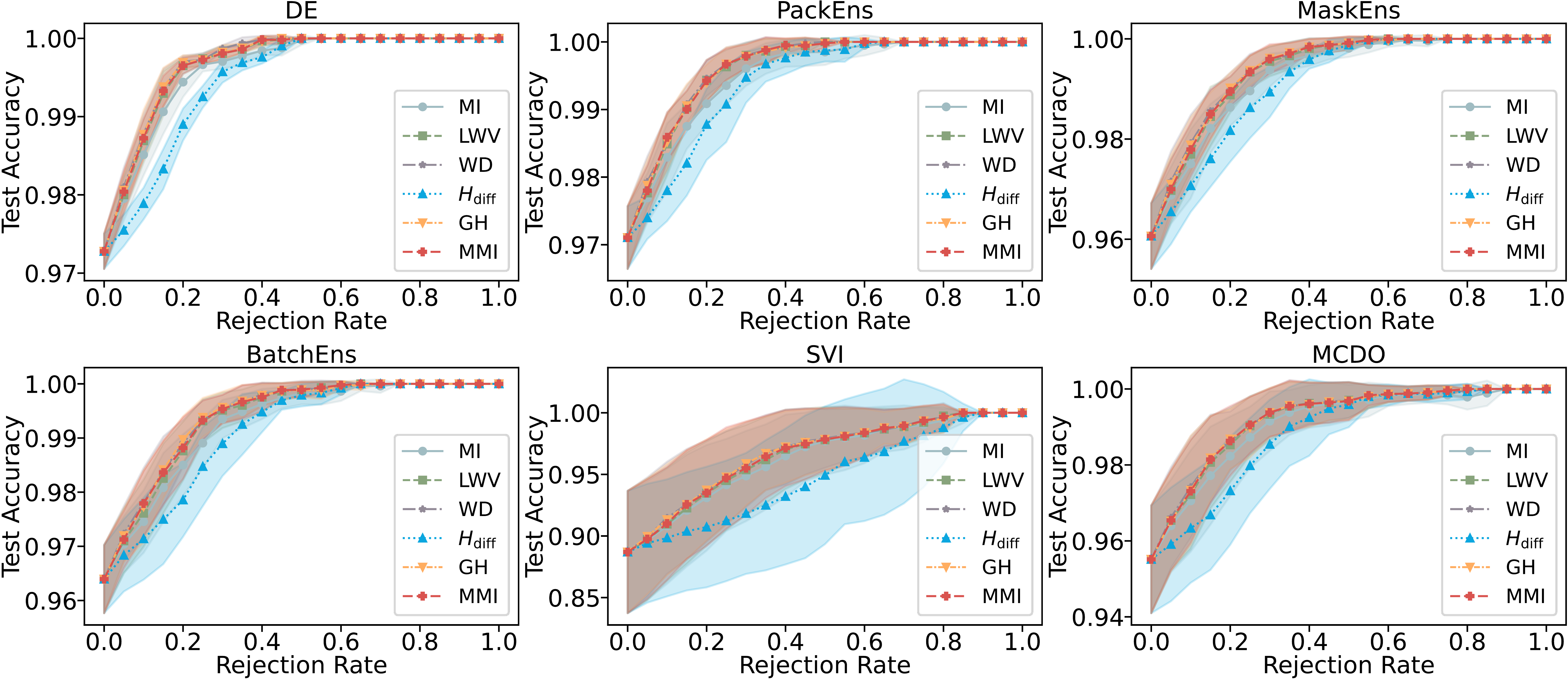}
\caption{Accuracy rejection curves using different uncertainty representations and measures on in-distribution SeaShip test data across different underlying predictive models.}
\label{FIG: ShipARC}
\end{figure}
\begin{figure}[!htbp]
\centering
\includegraphics[width=\linewidth]{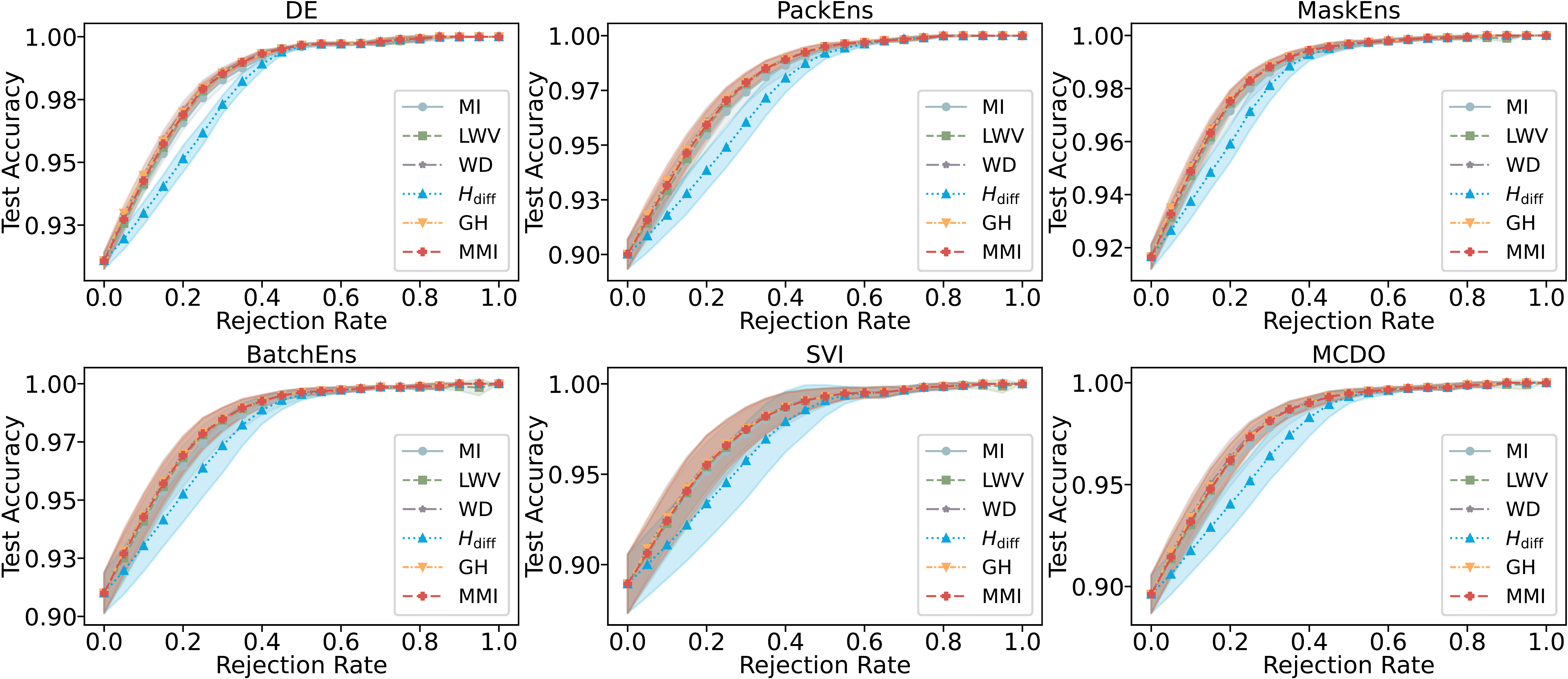}
\caption{Accuracy rejection curves using different uncertainty representations and measures on in-distribution CIFAR10 test data across different underlying predictive models.}
\label{FIG: CIFARARC}
\end{figure}

\begin{figure}[!htbp]
\centering
\includegraphics[width=\linewidth]{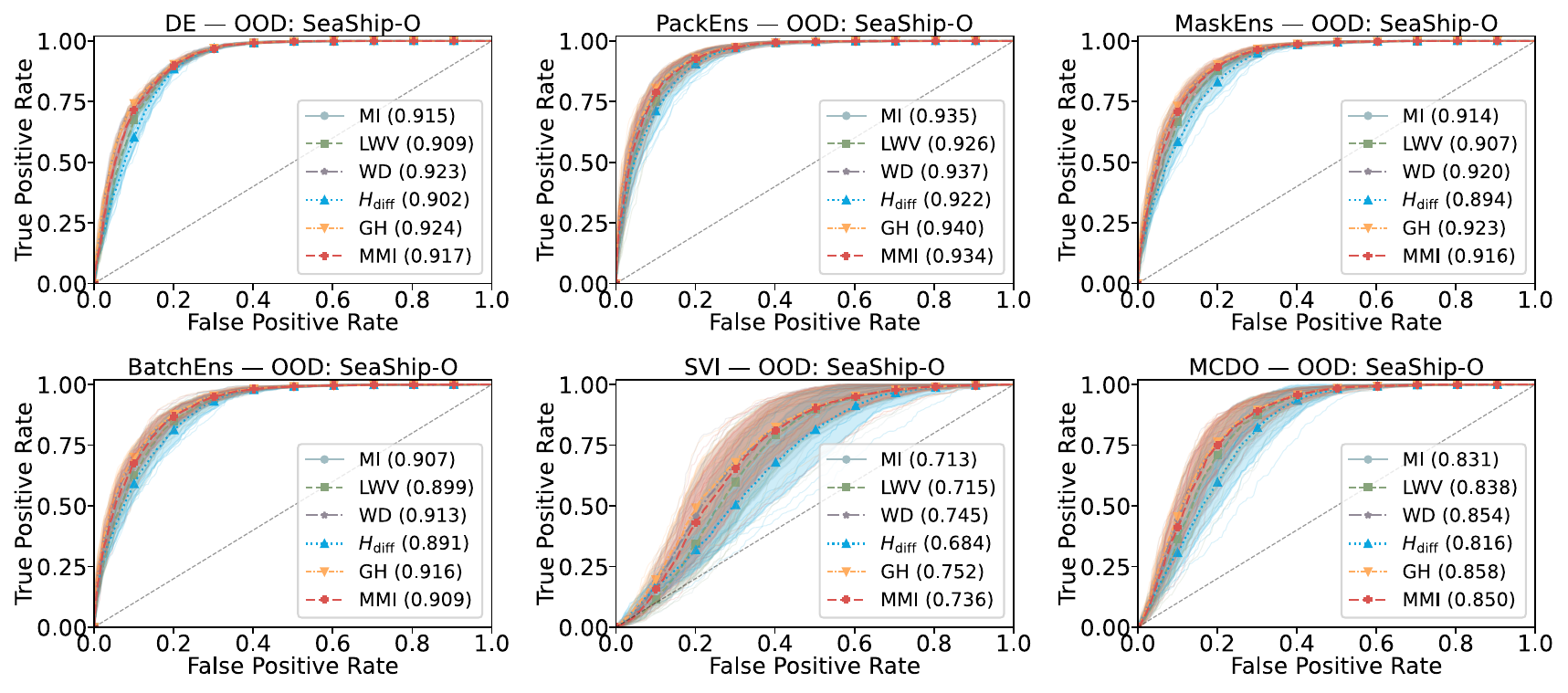}
\caption{ROC curves for OOD detection (SeaShip v.s. Seachip-O) across uncertainty measures and backbone methods. Each panel corresponds to one backbone, with all uncertainty measures plotted together. Solid lines represent the mean TPR over $10$ independent runs, shaded regions indicate $\pm1$ standard deviation, and faint lines show individual runs. Mean AUROC values are reported in the legend.}
\label{FIG: ROCShipO}
\end{figure}

\begin{figure}[!htbp]
\centering
\includegraphics[width=\linewidth]{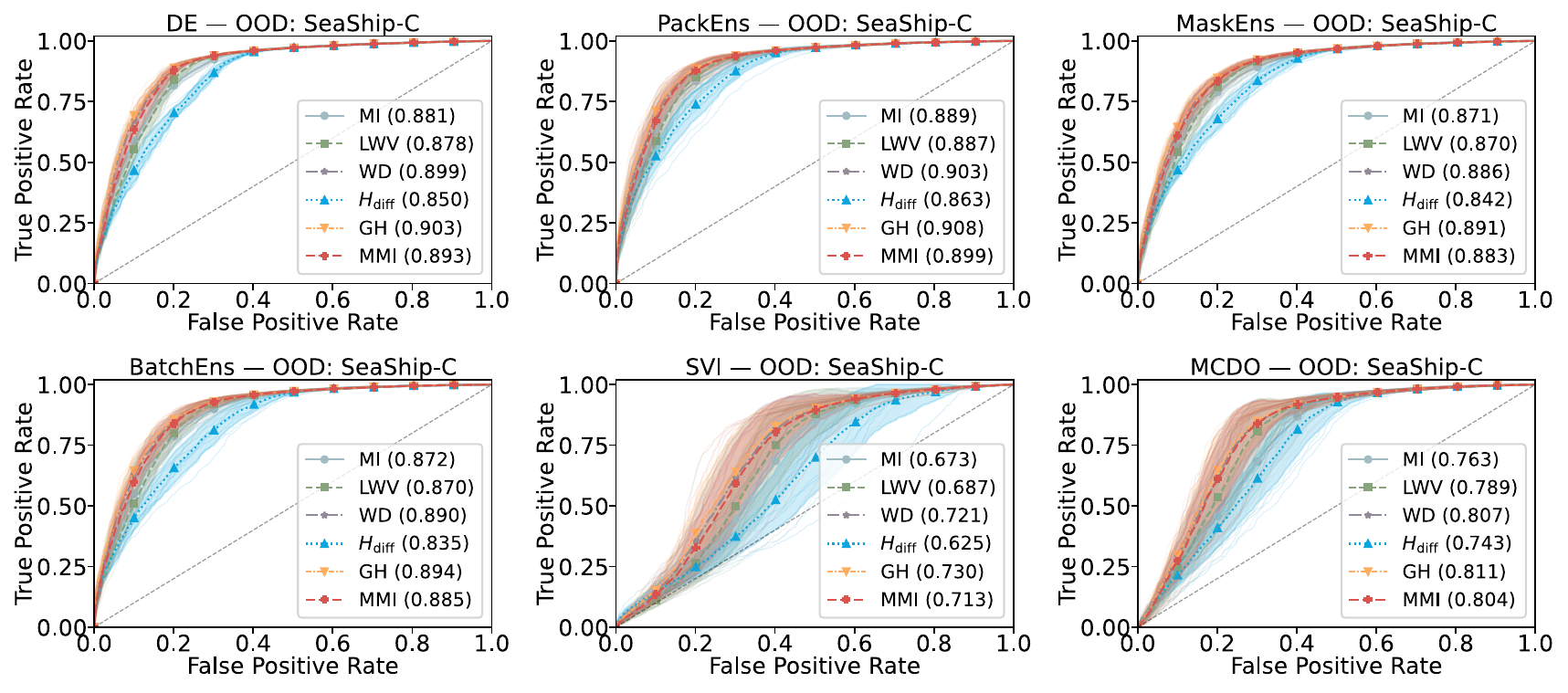}
\caption{ROC curves for OOD detection (SeaShip v.s. Seachip-C) across uncertainty measures and backbone methods. Each panel corresponds to one backbone, with all uncertainty measures plotted together. Solid lines represent the mean TPR over $10$ independent runs, shaded regions indicate $\pm1$ standard deviation, and faint lines show individual runs. Mean AUROC values are reported in the legend.}
\label{FIG: ROCShipC}
\end{figure}

\begin{figure}[!htbp]
\centering
\includegraphics[width=\linewidth]{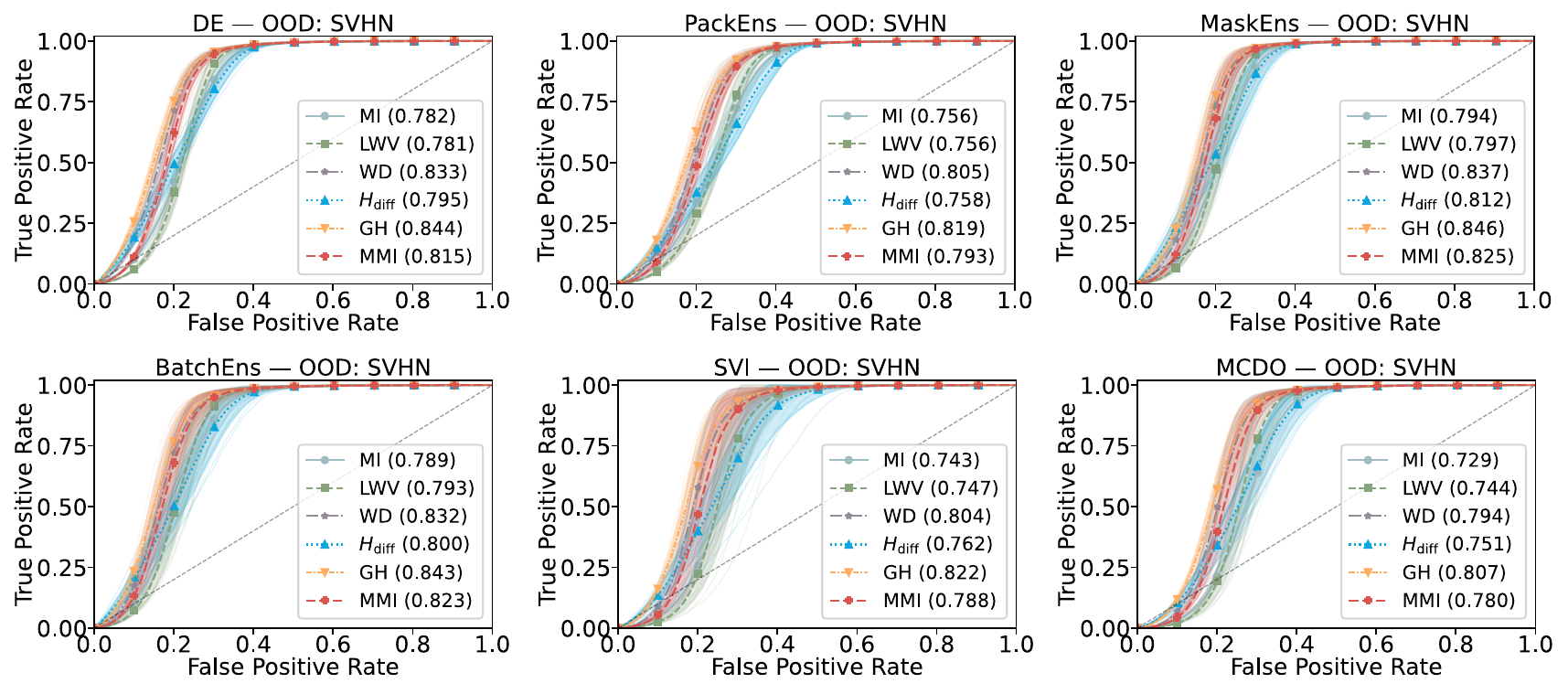}
\caption{ROC curves for OOD detection (CIFAR10 v.s. SVHN) across uncertainty measures and backbone methods. Each panel corresponds to one backbone, with all uncertainty measures plotted together. Solid lines represent the mean TPR over $10$ independent runs, shaded regions indicate $\pm1$ standard deviation, and faint lines show individual runs. Mean AUROC values are reported in the legend.}
\label{FIG: ROCSVHN}
\end{figure}

\begin{figure}[!htbp]
\centering
\includegraphics[width=\linewidth]{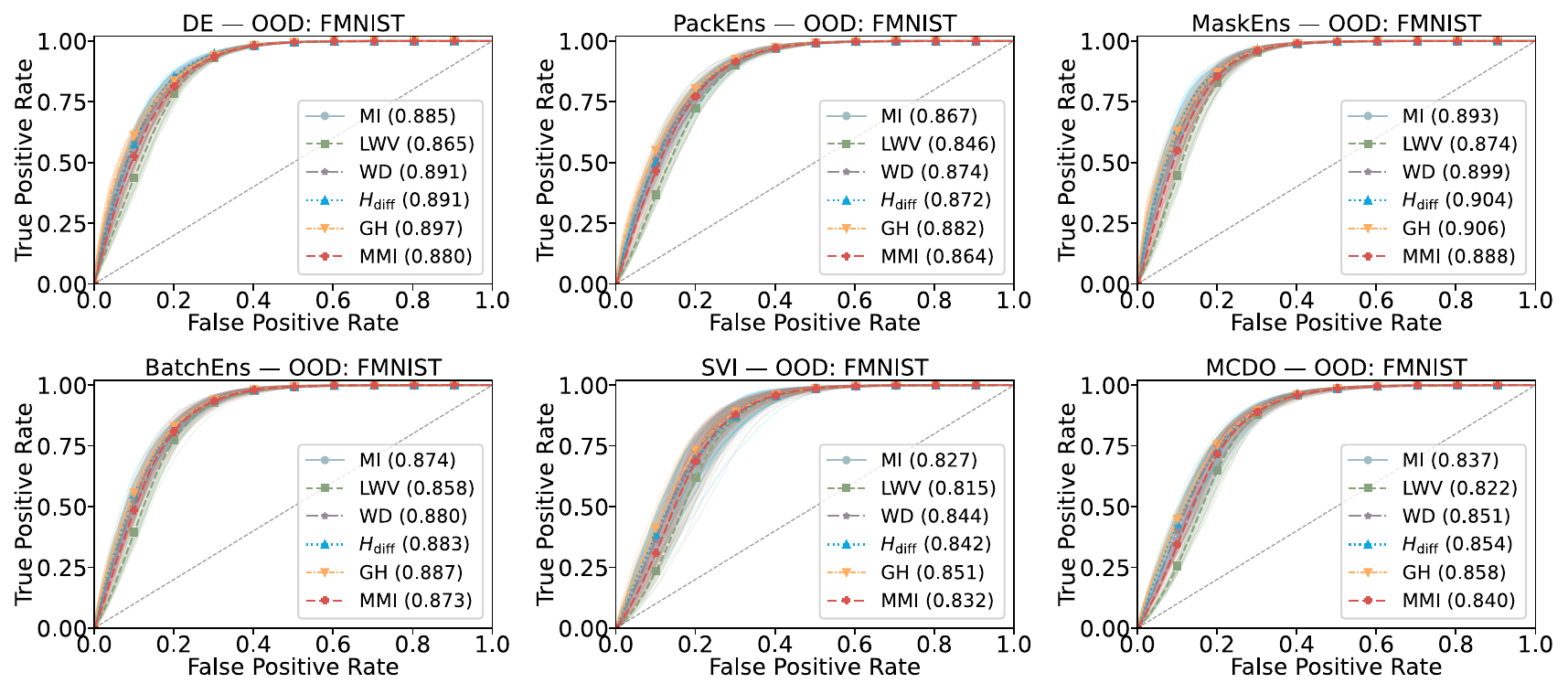}
\caption{ROC curves for OOD detection (CIFAR10 v.s. FMNIST) across uncertainty measures and backbone methods. Each panel corresponds to one backbone, with all uncertainty measures plotted together. Solid lines represent the mean TPR over $10$ independent runs, shaded regions indicate $\pm1$ standard deviation, and faint lines show individual runs. Mean AUROC values are reported in the legend.}
\label{FIG: ROCFMNIST}
\end{figure}
\newpage
\section{Further Experiments}
\label{App: furtherExps}
\begin{table*}[!htbp]
\caption{Average AUARC on the selective prediction task and AUROC on OOD detection tasks for models trained on \textbf{FMNIST}, averaged across six distinct predictive models. Intra- and inter-representation net wins are reported alongside. The first rank is shown in \textbf{\textcolor{red}{bold red}}, the second in \textbf{\textcolor{blue}{bold blue}}.}
\label{Table: FMNISTSummary}
\centering
\small
\setlength\tabcolsep{10pt}
\begin{tabular}{@{}lccc|ccc|ccc@{}}
\toprule
\multirow{2}{*}{}   & \multicolumn{3}{c|}{\textbf{Selective Prediction}}                                                & \multicolumn{3}{c|}{\textbf{OOD Detection (SVHN)}}                                             & \multicolumn{3}{c}{\textbf{OOD Detection (CIFAR10)}}                                                   \\ \cmidrule(l){2-10} 
                    & Average scores                                      & Intra                  & Inter & Average scores                                      & Intra                  & Inter & Average scores                                      & Intra                  & Inter \\ \midrule
MI                  & \multicolumn{1}{c|}{0.9901\scriptsize{$\pm$0.0017}} & \multicolumn{1}{c|}{-12}  & -18     & \multicolumn{1}{c|}{0.9148\scriptsize{$\pm$0.0713}} & \multicolumn{1}{c|}{\textcolor{blue}{\textbf{-1}}}  & -8     & \multicolumn{1}{c|}{0.9074\scriptsize{$\pm$0.0701}} & \multicolumn{1}{c|}{\textcolor{red}{\textbf{6}}}  & 3     \\
LWV                 & \multicolumn{1}{c|}{0.9910\scriptsize{$\pm$0.0016}} & \multicolumn{1}{c|}{\textcolor{blue}{\textbf{0}}} & -6     & \multicolumn{1}{c|}{0.9026\scriptsize{$\pm$0.0604}} & \multicolumn{1}{c|}{-9} & -27     & \multicolumn{1}{c|}{0.8943\scriptsize{$\pm$0.0609}} & \multicolumn{1}{c|}{-10} & -28     \\
WD                  & \multicolumn{1}{c|}{0.9912\scriptsize{$\pm$0.0015}} & \multicolumn{1}{c|}{\textcolor{red}{\textbf{12}}} & \textcolor{red}{\textbf{25}}    & \multicolumn{1}{c|}{0.9323\scriptsize{$\pm$0.0511}} & \multicolumn{1}{c|}{\textcolor{red}{\textbf{10}}} & \textcolor{blue}{\textbf{14}}     & \multicolumn{1}{c|}{0.9126\scriptsize{$\pm$0.0536}} & \multicolumn{1}{c|}{\textcolor{blue}{\textbf{4}}} & -2     \\ \cmidrule(l){3-3}\cmidrule(lr){6-6}\cmidrule(lr){9-9}
GH                  & \multicolumn{1}{c|}{0.9912\scriptsize{$\pm$0.0015}} & \multicolumn{1}{c|}{\textcolor{red}{\textbf{12}}} & \textcolor{blue}{\textbf{23}}    & \multicolumn{1}{c|}{0.9414\scriptsize{$\pm$0.0473}} & \multicolumn{1}{c|}{\textcolor{red}{\textbf{12}}} & \textcolor{red}{\textbf{30}}    & \multicolumn{1}{c|}{0.9173\scriptsize{$\pm$0.0523}} & \multicolumn{1}{c|}{\textcolor{blue}{\textbf{1}}} & \textcolor{blue}{\textbf{12}}     \\
$H_{\text{diff}}$   & \multicolumn{1}{c|}{0.9874\scriptsize{$\pm$0.0026}} & \multicolumn{1}{c|}{-12}  & -30     & \multicolumn{1}{c|}{0.9215\scriptsize{$\pm$0.0612}} & \multicolumn{1}{c|}{-7}  & -4      & \multicolumn{1}{c|}{0.9306\scriptsize{$\pm$0.0554}} & \multicolumn{1}{c|}{\textcolor{red}{\textbf{11}}}  & \textcolor{red}{\textbf{29}}      \\
MMI                 & \multicolumn{1}{c|}{0.9911\scriptsize{$\pm$0.0015}} & \multicolumn{1}{c|}{\textcolor{blue}{\textbf{0}}} & 6    & \multicolumn{1}{c|}{0.9272\scriptsize{$\pm$0.0525}} & \multicolumn{1}{c|}{\textcolor{blue}{\textbf{-5}}} & -5     & \multicolumn{1}{c|}{0.9094\scriptsize{$\pm$0.0551}} & \multicolumn{1}{c|}{-12} & -14     \\ \bottomrule
\end{tabular}
\end{table*}

\begin{figure}[!htbp]
\centering
\includegraphics[width=\linewidth]{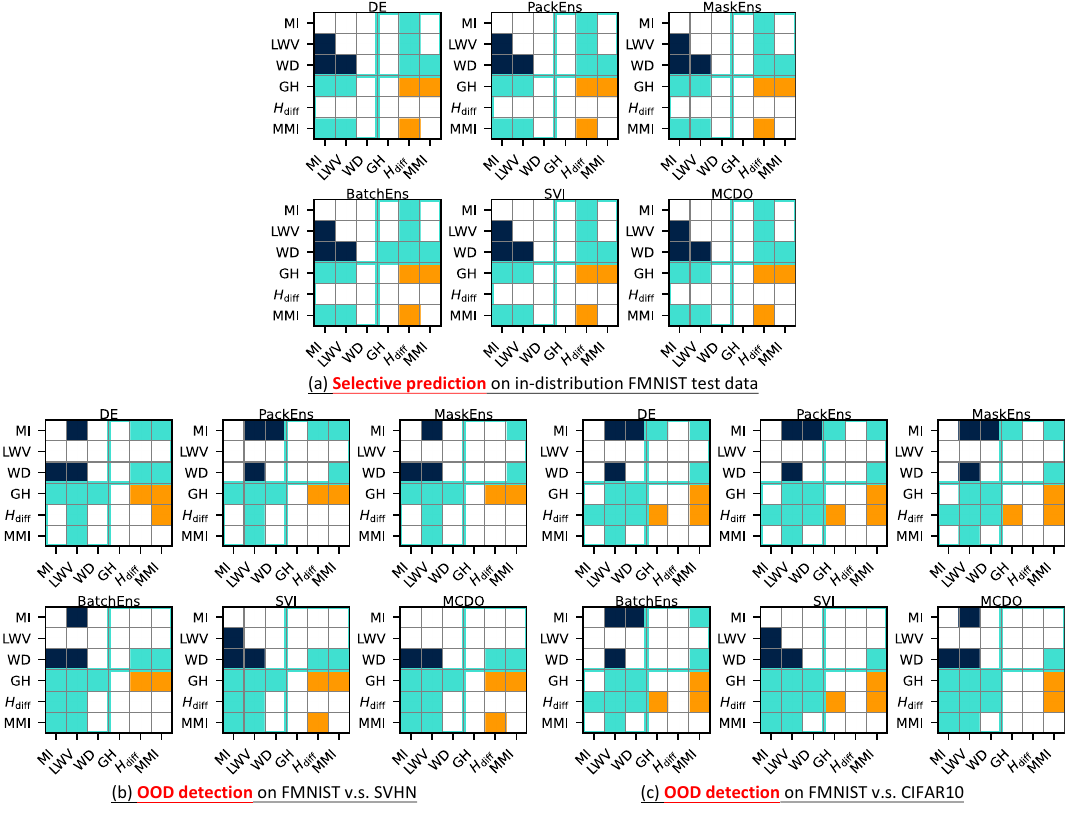}
\caption{Statistical significance plots on different selective prediction (a) and OOD detection (b-c) benchmarks across different underlying predictive models trained on the \textbf{FMNIST} dataset. A cell is shaded if the measure in the $i$-th row is statistically significantly better than that in the $j$-th column according to a pairwise one-sided Wilcoxon signed-rank test at the 5\% significance level. Intra-representation comparisons are shown in blue (distribution-based measures) and orange (credal-based measures), while inter-representation comparisons are shown in green.}
\label{FIG: FMNISTresults}
\end{figure}

\begin{table*}[!htbp]
\caption{Average AUARC on the selective prediction task and AUROC on OOD detection tasks for models trained on \textbf{SVHN}, averaged across six distinct predictive models. Intra- and inter-representation net wins are reported alongside. The first rank is shown in \textbf{\textcolor{red}{bold red}}, the second in \textbf{\textcolor{blue}{bold blue}}.}
\label{Table: SVHNSummary}
\centering
\small
\setlength\tabcolsep{10pt}
\begin{tabular}{@{}lccc|ccc|ccc@{}}
\toprule
\multirow{2}{*}{} & \multicolumn{3}{c|}{\textbf{Selective Prediction}} & \multicolumn{3}{c|}{\textbf{OOD Detection (FMNIST)}} & \multicolumn{3}{c}{\textbf{OOD Detection (CIFAR10)}} \\ \cmidrule(l){2-10}
 & Average scores & Intra & Inter & Average scores & Intra & Inter & Average scores & Intra & Inter \\ \midrule
MI
 & \multicolumn{1}{c|}{0.9950\scriptsize{$\pm$0.0007}}
 & \multicolumn{1}{c|}{-10}
 & -16
 & \multicolumn{1}{c|}{0.9704\scriptsize{$\pm$0.0108}}
 & \multicolumn{1}{c|}{\textcolor{red}{\textbf{6}}}
 & \textcolor{blue}{\textbf{13}}
 & \multicolumn{1}{c|}{0.9765\scriptsize{$\pm$0.0074}}
 & \multicolumn{1}{c|}{\textcolor{red}{\textbf{6}}}
 & \textcolor{blue}{\textbf{12}} \\
LWV
 & \multicolumn{1}{c|}{0.9952\scriptsize{$\pm$0.0006}}
 & \multicolumn{1}{c|}{\textcolor{blue}{\textbf{2}}}
 & 5
 & \multicolumn{1}{c|}{0.9570\scriptsize{$\pm$0.0122}}
 & \multicolumn{1}{c|}{-12}
 & -30
 & \multicolumn{1}{c|}{0.9629\scriptsize{$\pm$0.0097}}
 & \multicolumn{1}{c|}{-12}
 & -28 \\
WD
 & \multicolumn{1}{c|}{0.9953\scriptsize{$\pm$0.0007}}
 & \multicolumn{1}{c|}{\textcolor{red}{\textbf{8}}}
 & \textcolor{red}{\textbf{22}}
 & \multicolumn{1}{c|}{0.9704\scriptsize{$\pm$0.0092}}
 & \multicolumn{1}{c|}{\textcolor{red}{\textbf{6}}}
 & 12
 & \multicolumn{1}{c|}{0.9769\scriptsize{$\pm$0.0062}}
 & \multicolumn{1}{c|}{\textcolor{red}{\textbf{6}}}
 & \textcolor{blue}{\textbf{12}} \\ \cmidrule(l){3-3}\cmidrule(lr){6-6}\cmidrule(lr){9-9}
GH
 & \multicolumn{1}{c|}{0.9953\scriptsize{$\pm$0.0007}}
 & \multicolumn{1}{c|}{\textcolor{red}{\textbf{10}}}
 & \textcolor{blue}{\textbf{15}}
 & \multicolumn{1}{c|}{0.9741\scriptsize{$\pm$0.0082}}
 & \multicolumn{1}{c|}{\textcolor{red}{\textbf{12}}}
 & \textcolor{red}{\textbf{29}}
 & \multicolumn{1}{c|}{0.9806\scriptsize{$\pm$0.0051}}
 & \multicolumn{1}{c|}{\textcolor{red}{\textbf{12}}}
 & \textcolor{red}{\textbf{30}} \\
$H_{\text{diff}}$
 & \multicolumn{1}{c|}{0.9944\scriptsize{$\pm$0.0010}}
 & \multicolumn{1}{c|}{-12}
 & -30
 & \multicolumn{1}{c|}{0.9649\scriptsize{$\pm$0.0118}}
 & \multicolumn{1}{c|}{-9}
 & -15
 & \multicolumn{1}{c|}{0.9653\scriptsize{$\pm$0.0100}}
 & \multicolumn{1}{c|}{-12}
 & -20 \\
MMI
 & \multicolumn{1}{c|}{0.9953\scriptsize{$\pm$0.0007}}
 & \multicolumn{1}{c|}{\textcolor{blue}{\textbf{2}}}
 & 4
 & \multicolumn{1}{c|}{0.9665\scriptsize{$\pm$0.0100}}
 & \multicolumn{1}{c|}{\textcolor{blue}{\textbf{-3}}}
 & -9
 & \multicolumn{1}{c|}{0.9730\scriptsize{$\pm$0.0072}}
 & \multicolumn{1}{c|}{\textcolor{blue}{\textbf{0}}}
 & -6 \\ \bottomrule
\end{tabular}
\end{table*}

\begin{figure}[!htbp]
\centering
\includegraphics[width=\linewidth]{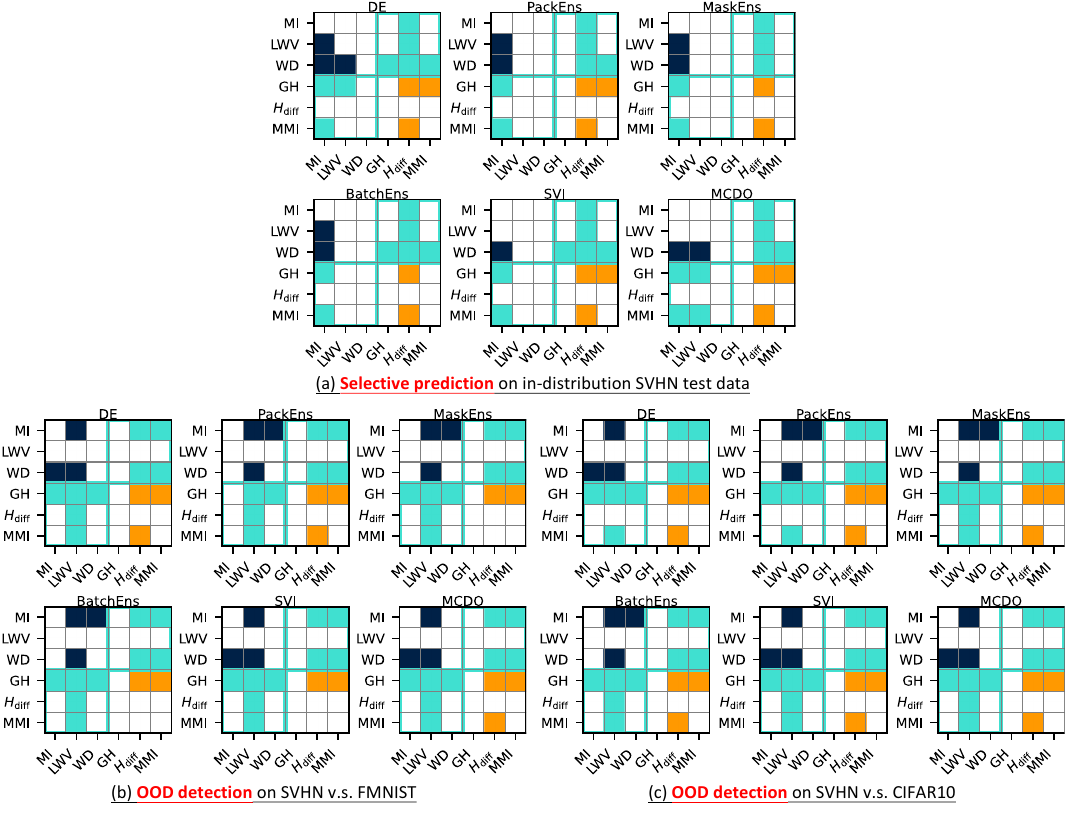}
\caption{Statistical significance plots on different selective prediction (a) and OOD detection (b-c) benchmarks across different underlying predictive models trained on the \textbf{SVHN} dataset. A cell is shaded if the measure in the $i$-th row is statistically significantly better than that in the $j$-th column according to a pairwise one-sided Wilcoxon signed-rank test at the 5\% significance level. Intra-representation comparisons are shown in blue (distribution-based measures) and orange (credal-based measures), while inter-representation comparisons are shown in green.}
\label{FIG: SVHNresults}
\end{figure}

\begin{figure}[!htbp]
\centering
\includegraphics[width=\linewidth]{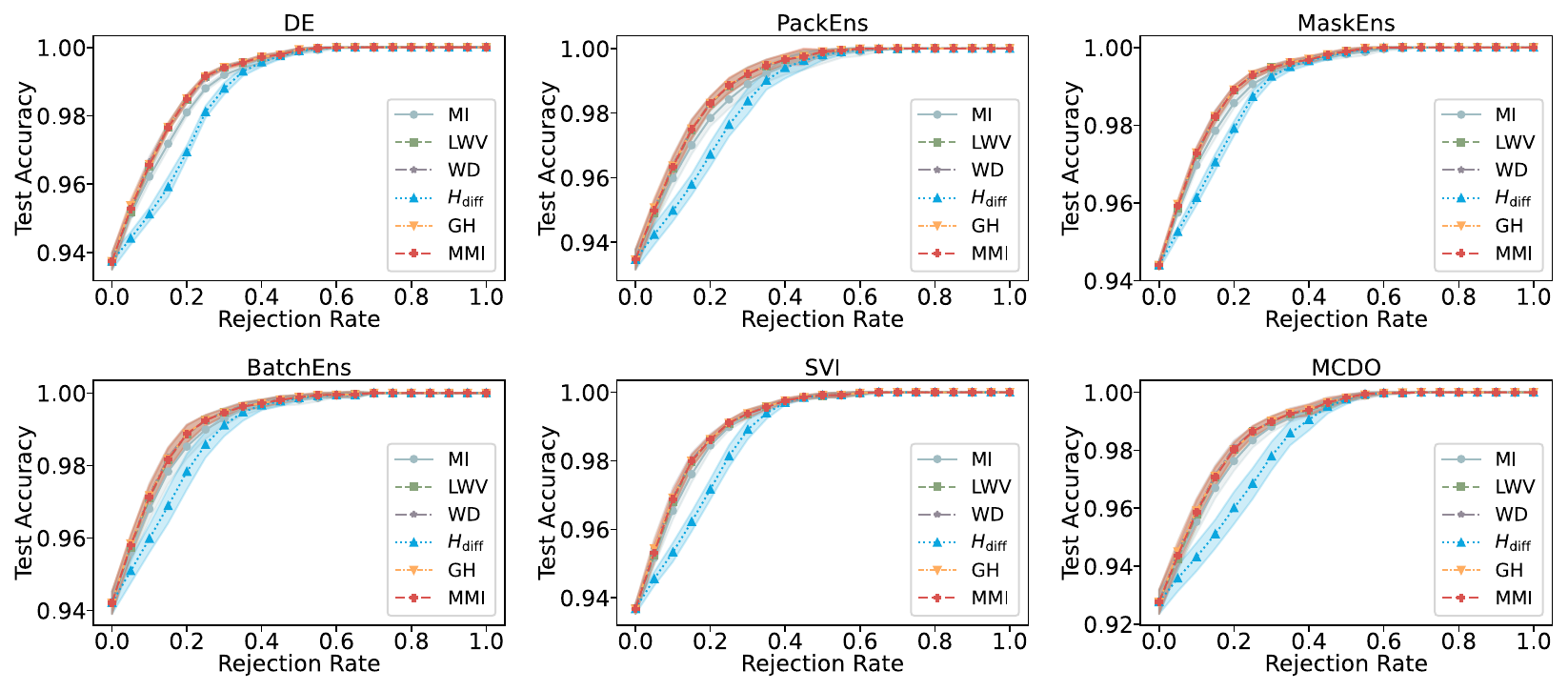}
\caption{Accuracy rejection curves using different uncertainty representations and measures on in-distribution \textbf{FMNIST} test data across different underlying predictive models.}
\label{FIG: FMNISTARC}
\end{figure}

\begin{figure}[!htbp]
\centering
\includegraphics[width=\linewidth]{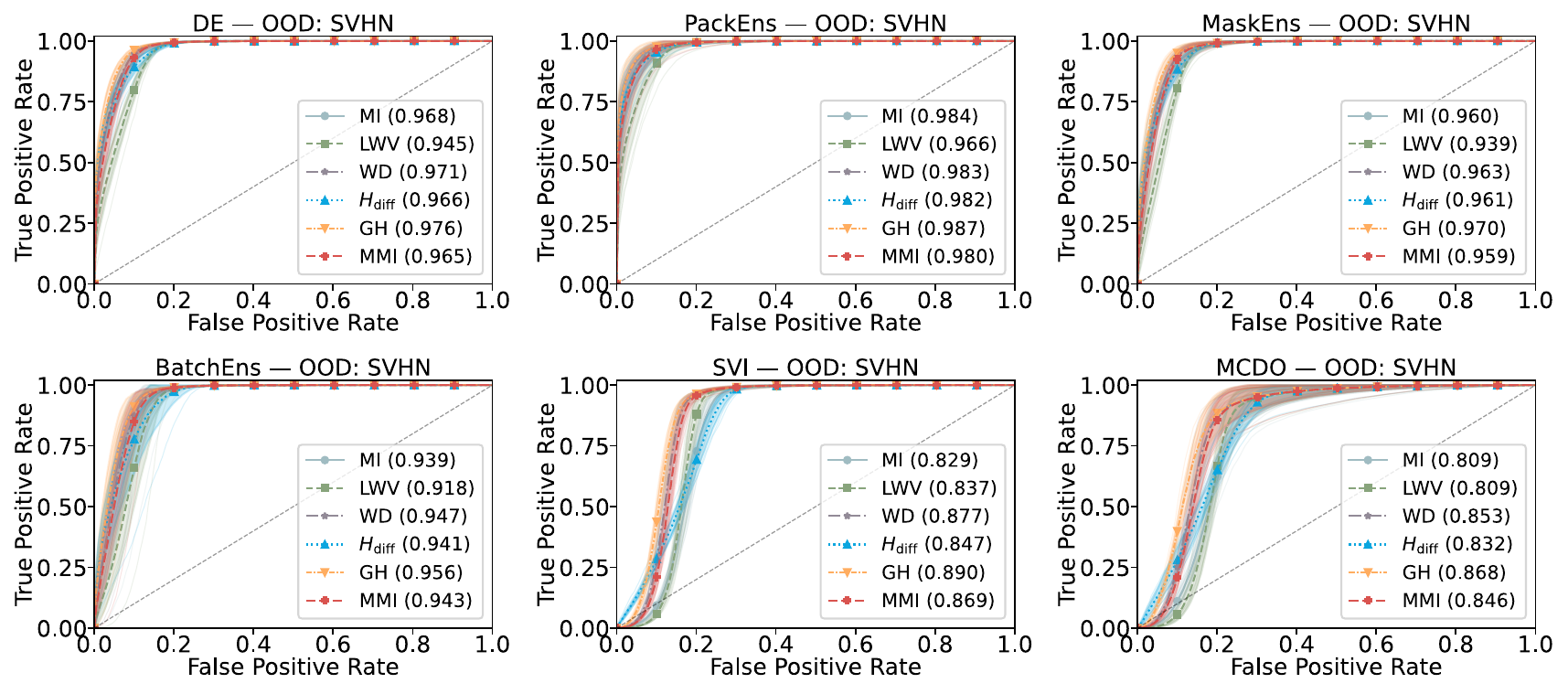}
\caption{ROC curves for OOD detection (\textbf{ID: FMNIST v.s. OOD: SVHN}) across uncertainty measures and backbone methods. Each panel corresponds to one backbone, with all uncertainty measures plotted together. Solid lines represent the mean TPR over $10$ independent runs, shaded regions indicate $\pm1$ standard deviation, and faint lines show individual runs. Mean AUROC values are reported in the legend.}
\label{FIG: ROCFMNISTvsSVHN}
\end{figure}
\begin{figure}[!htbp]
\centering
\includegraphics[width=\linewidth]{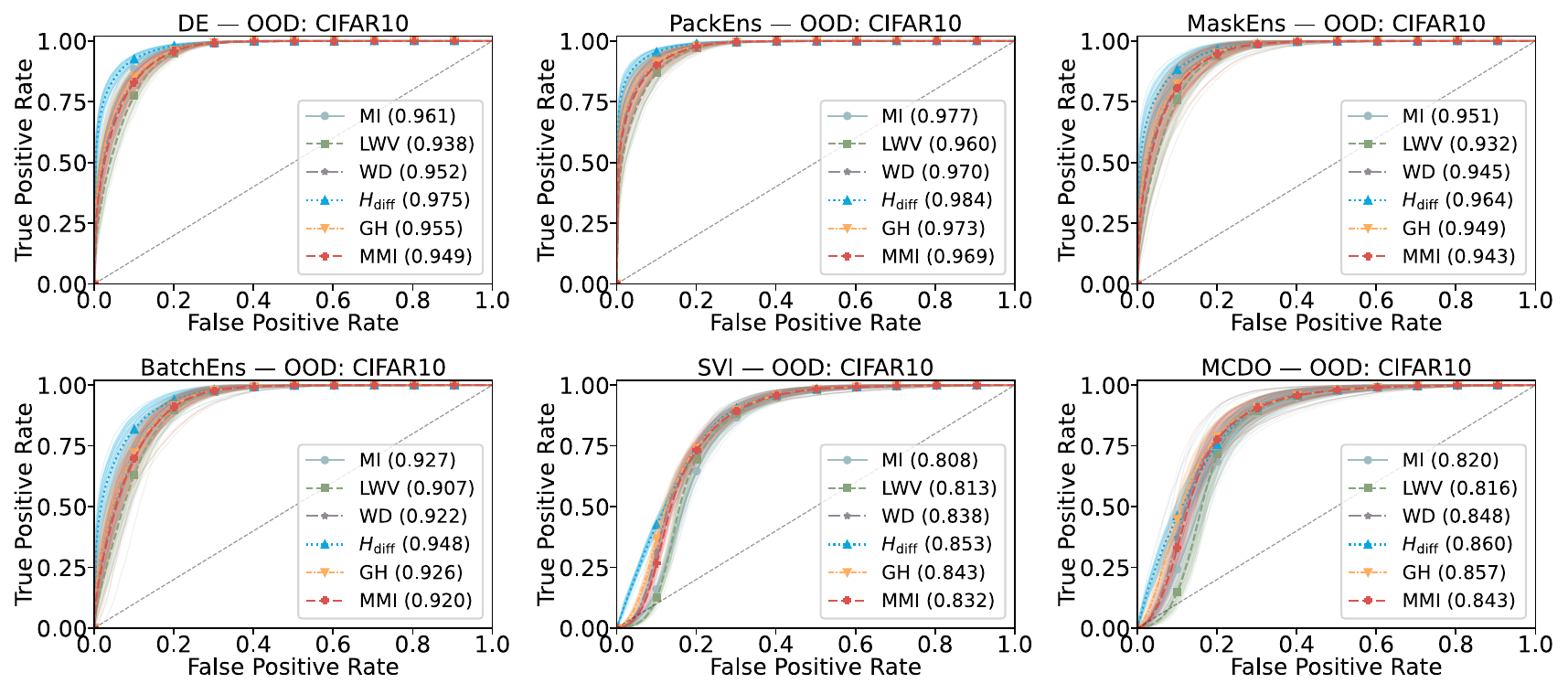}
\caption{ROC curves for OOD detection (\textbf{ID: FMNIST v.s. OOD: CIFAR10}) across uncertainty measures and backbone methods. Each panel corresponds to one backbone, with all uncertainty measures plotted together. Solid lines represent the mean TPR over $10$ independent runs, shaded regions indicate $\pm1$ standard deviation, and faint lines show individual runs. Mean AUROC values are reported in the legend.}
\label{FIG: ROCFMNISTvsCIFAR}
\end{figure}

\begin{figure}[!htbp]
\centering
\includegraphics[width=\linewidth]{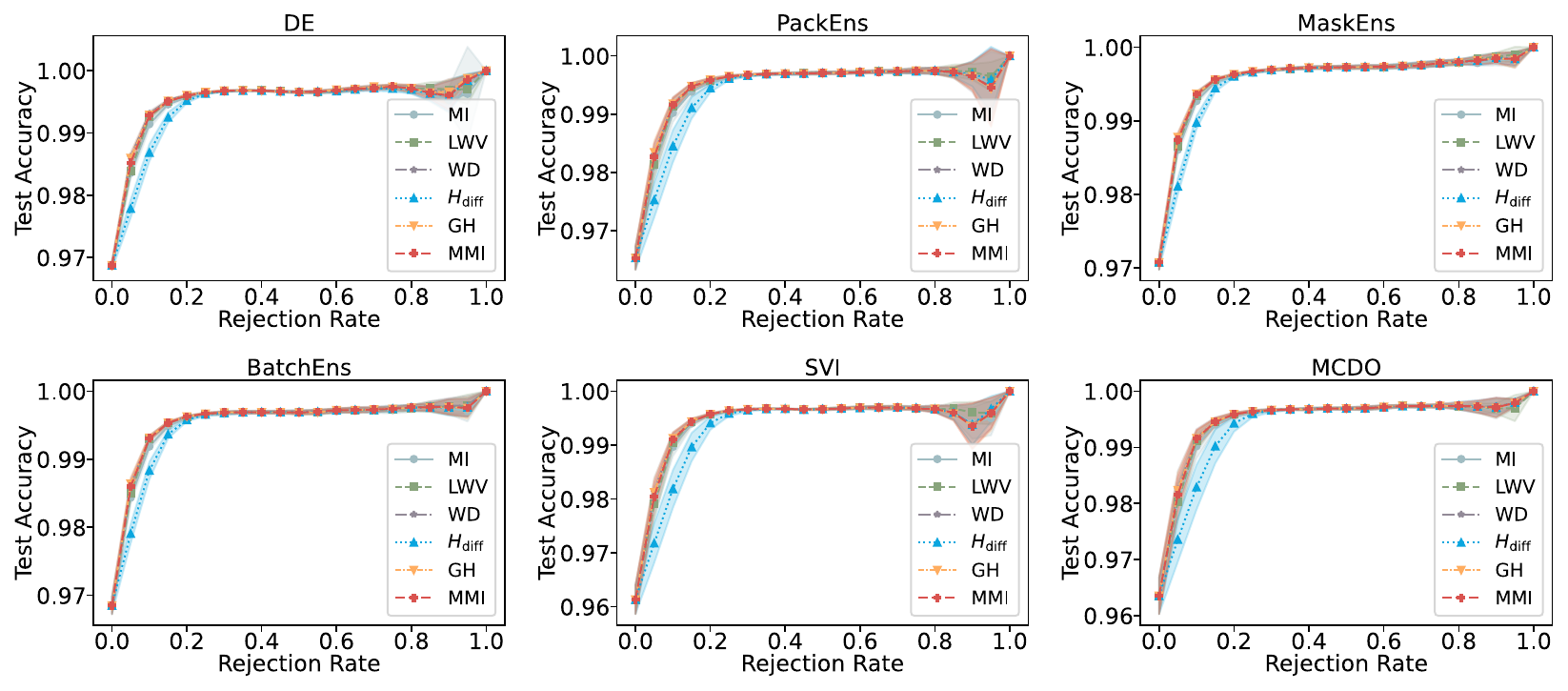}
\caption{Accuracy rejection curves using different uncertainty representations and measures on in-distribution \textbf{SVHN} test data across different underlying predictive models.}
\label{FIG: SVHNARC}
\end{figure}

\begin{figure}[!htbp]
\centering
\includegraphics[width=\linewidth]{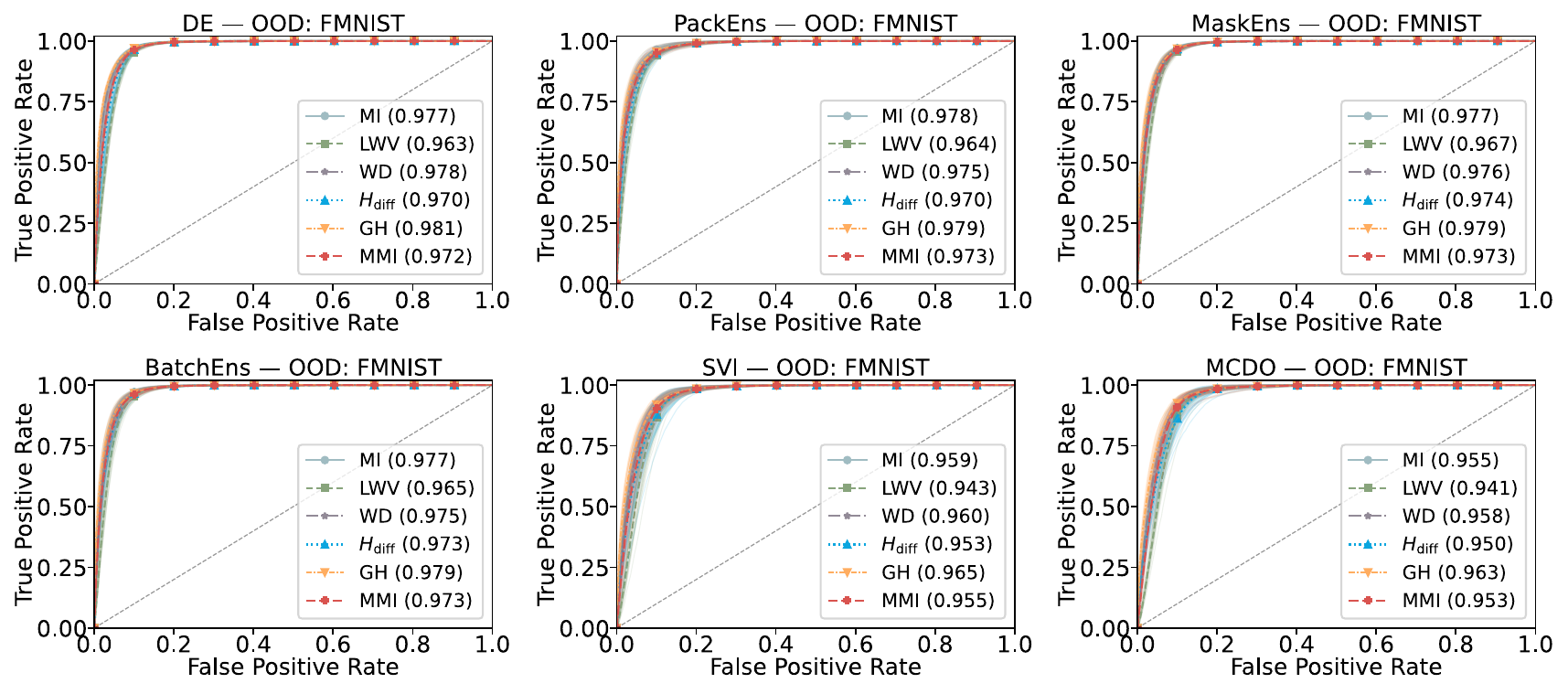}
\caption{ROC curves for OOD detection (\textbf{ID: SVHN v.s. OOD: FMNIST}) across uncertainty measures and backbone methods. Each panel corresponds to one backbone, with all uncertainty measures plotted together. Solid lines represent the mean TPR over $10$ independent runs, shaded regions indicate $\pm1$ standard deviation, and faint lines show individual runs. Mean AUROC values are reported in the legend.}
\label{FIG: ROCSVHNvsFMNIST}
\end{figure}
\begin{figure}[!htbp]
\centering
\includegraphics[width=\linewidth]{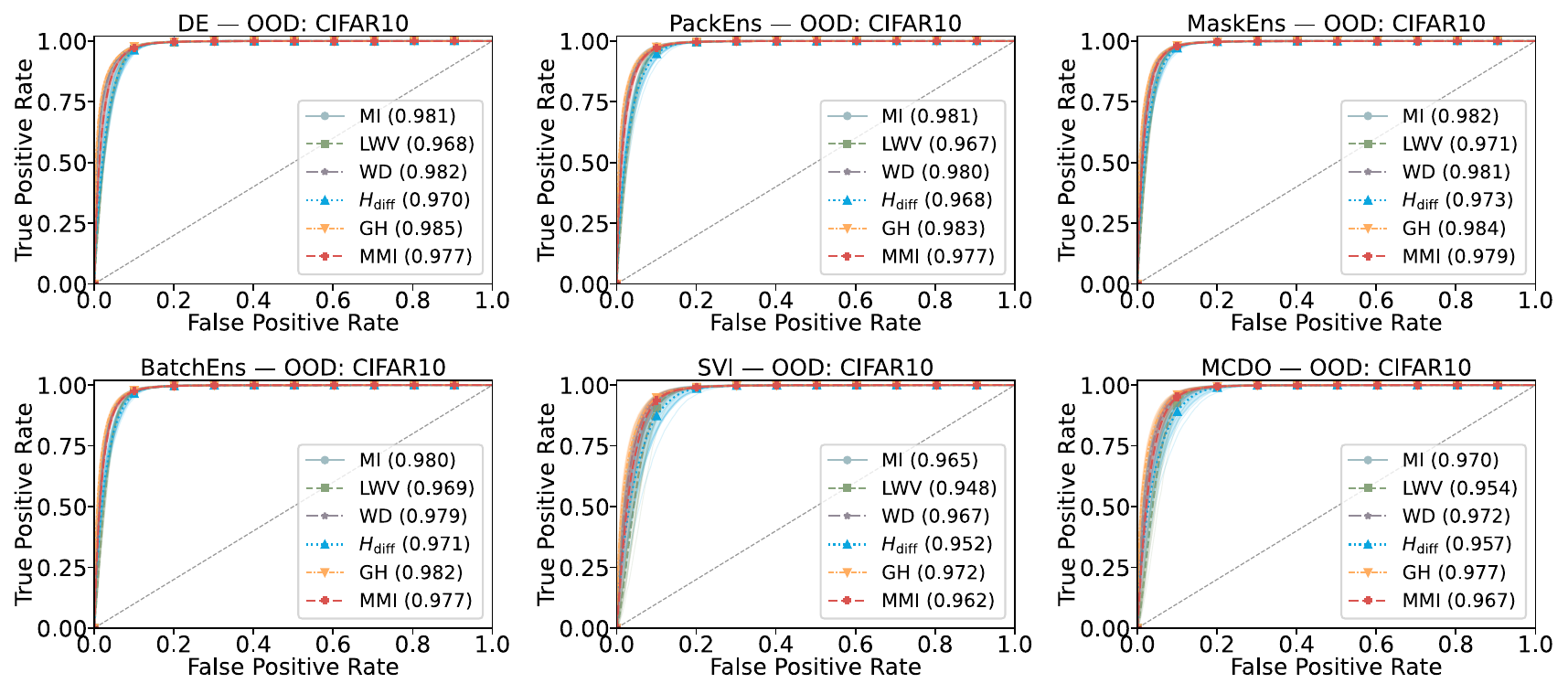}
\caption{ROC curves for OOD detection (\textbf{ID: SVHN v.s. OOD: CIFAR10}) across uncertainty measures and backbone methods. Each panel corresponds to one backbone, with all uncertainty measures plotted together. Solid lines represent the mean TPR over $10$ independent runs, shaded regions indicate $\pm1$ standard deviation, and faint lines show individual runs. Mean AUROC values are reported in the legend.}
\label{FIG: ROCSVHNvsCIFAR}
\end{figure}
\end{document}